\pdfoutput=1

\documentclass[11pt]{article}

\usepackage[]{EMNLP2023}

\usepackage{times}
\usepackage{latexsym}

\usepackage[T1]{fontenc}

\usepackage[utf8]{inputenc}
\usepackage{CJKutf8}

\usepackage{microtype}

\usepackage{inconsolata}

\usepackage{graphicx}
\usepackage{tikz}
\usepackage{tabularx}
\usepackage{amsmath}
\usepackage{subcaption}
\usepackage{placeins}
\usepackage{float}
\usepackage[export]{adjustbox}

\title{The Impact of Familiarity on Naming Variation: A Study on Object Naming in Mandarin Chinese}

\author{Yunke He$^\textbf{1}$, Xixian Liao$^\textbf{1}$, Jialing Liang$^\textbf{1}$ \and Gemma Boleda$^\textbf{1,2}$ \\
$^1$Department of Translation and Language Sciences, Universitat Pompeu Fabra \\
$^2$Catalan Institution for Research and Advanced Studies - ICREA\\
\texttt {\{xixian.liao,jialing.liang,gemma.boleda\}@upf.edu} \\
\texttt {yunkehe66@gmail.com}\\}

\begin{document}
\maketitle
\begin{abstract}
Different speakers often produce different names for the same object or entity (e.g., ``woman'' vs.\ ``tourist'' for a female tourist). 
The reasons behind variation in naming are not well understood.
We create a Language and Vision dataset for Mandarin Chinese that provides an average of 20 names for 1319 naturalistic images, and investigate how familiarity with a given kind of object relates to the degree of naming variation it triggers across subjects.
We propose that familiarity influences naming variation in two competing ways: increasing familiarity can either expand vocabulary, leading to higher variation, or promote convergence on conventional names, thereby reducing variation.
We find evidence for both factors being at play.
Our study illustrates how computational resources can be used to address research questions in Cognitive Science.
\end{abstract}

\section{Introduction}


\begin{CJK}{UTF8}{gbsn}
\begin{figure*}[htbp]
  \begin{subfigure}[b]{0.3\linewidth}
    \raisebox{0pt}[\height][0pt]{\includegraphics[width=\linewidth]{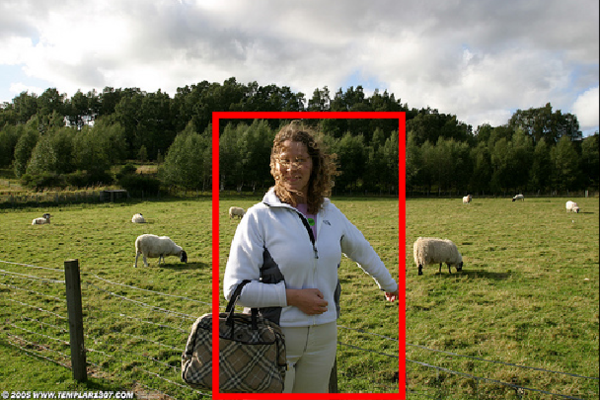}}
    \begin{minipage}{\textwidth}
        \raggedright
        \small
        女人 (12), 女士 (2), 人 (2), 大人 (1), 女 (1), 游客 (1) \\
        woman (12), lady (2), person (2), adult (1), female (1), tourist (1) \\
        \textbf{Familiarity: 4.2} / \textbf{H: 1.8} / \textbf{N: 6} \\
    \end{minipage}  
    \caption{}
    \label{fig:woman}
  \end{subfigure}\hfill
  \begin{subfigure}[b]{0.3\linewidth}
    \raisebox{0pt}[\height][0pt]{\includegraphics[width=\linewidth]{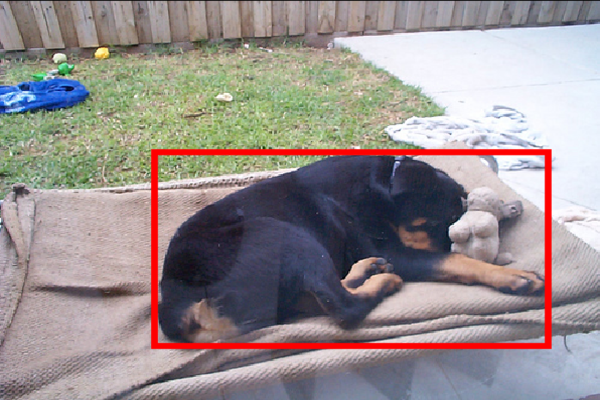}}
    \begin{minipage}{\textwidth}
        \raggedright
        \small
        狗 (21), 狗狗 (1),罗威勒狗 (1) \\
        \vspace{3.1mm}
        dog (21), puppy (1), Rottweiler (1) \\
        \vspace{4mm}
        \textbf{Familiarity: 4.1} / \textbf{H: 0.5} /  \textbf{N: 3}\\
    \end{minipage}
    \caption{}
    \label{fig:dog}
  \end{subfigure}
  \hfill
  \begin{subfigure}[b]{0.3\linewidth}
    \raisebox{0pt}[\height][0pt]{\includegraphics[width=\linewidth]{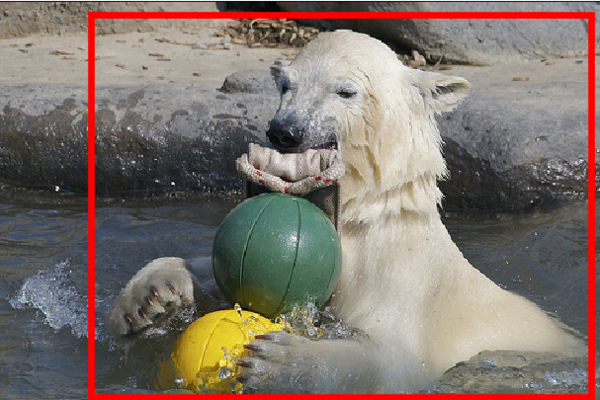}}
    \begin{minipage}{\textwidth}
        \raggedright
        \small
        北极熊 (8), 熊 (7), 动物 (2), 狗 (1), 海马 (1), 杂技 (1) \\ 
        polar bear (8), bear (7), animal (2), dog (1), seahorse (1), acrobatics (1)\\
        \textbf{Familiarity: 2.5} / \textbf{H: 2} /  \textbf{N: 6} \\
    \end{minipage}
    \caption{}
    \label{fig:polar bear}
  \end{subfigure}
  \caption{Examples of images and their corresponding names in ManyNames ZH. Numbers in parentheses are counts across subjects. Familiarity is estimated by weighted average of lexical frequency (see section \ref{sec:analysis-method}); H, or entropy, measures naming variation (see section \ref{sec:analysis-method}); N is the number of distinct names.}
  \label{fig:combined}
\end{figure*}
\end{CJK} 

When talking about objects in everyday experiences, people need to engage in the cognitive process of searching their lexicon to identify the most appropriate name to refer to them. This process involves intricate cognitive mechanisms that enable us to connect the properties of the object with the corresponding entries in our lexicon. Often, different individuals use different names to refer to the same object, reflecting the inherent variability in how we categorize and label our surroundings \cite{brown1958shall}; for instance, the woman in Figure \ref{fig:woman} can be called ``woman'', ``tourist'', or ``person'', among other choices.
The reasons behind this variability are still not well understood. 

Most previous research on naming has been done in Western languages (mostly English); and, in Cognitive Science, mostly with highly idealized stimuli, such as drawings of prototypical objects for a given category. \citet{silberer2020humans,silberer2020object} introduced ManyNames, a dataset with realistic stimuli which provides an average of 31 English names for 25K objects in naturalistic images such as those in Figure \ref{fig:combined}. 
In this study, we present ManyNames ZH,%
\footnote{Available at \url{https://github.com/flyingpiggy1214/ManyNames_ZH}} 
a new dataset for object naming that provides Mandarin Chinese names for a subset of the ManyNames data (1319 images, average 20 names per image).
Figure \ref{fig:combined} shows three example images with their corresponding names in ManyNames ZH. 

We use this Language and Vision resource to address an open research question in Cognitive Science, namely, the role of object familiarity on naming variation. 
Familiarity is defined in psycholinguistic research as the level of prior exposure or knowledge that individuals have about specific stimuli, such as words and objects \citep{snodgrass1980standardized,anaki2009familiarity}.
We explore two seemingly opposite hypotheses, which respectively focus on two different aspects of naming variation: convergence on a conventional name, and size of the available vocabulary. 

\begin{CJK}{UTF8}{gbsn}
\textbf{Hypothesis 1} (H1) posits that higher familiarity results in lower variation. This is based on the assumption that people tend to converge on a conventional name for familiar objects. Conversely, less familiar kinds of objects afford different conceptualizations, potentially increasing naming variation. For instance, most people are arguably more familiar with dogs than with bears, and indeed in Figure \ref{fig:dog} Chinese subjects mostly converge on the majority name "狗" (``dog''), while they use a wider range of words to refer to the polar bear in Figure \ref{fig:polar bear}. H1 has received support in some, but not all studies in Cognitive Science (see Section~\ref{sec:background}).
\end{CJK}

\textbf{Hypothesis 2} (H2) instead suggests that higher familiarity is associated with increased naming variation. H2 is based on the idea that we need a larger vocabulary to refer to kinds of objects that we talk a lot about, to encode finer-grained distinctions in an efficient way \cite{gatewood1984familiarity}. 
For instance, \citet{silberer2020humans} note that people elicit more variation than animals in ManyNames; according to H2, this would be due to the availability of a varied lexicon covering different dimensions that are relevant to categorize people, such as age (``child''), gender (``woman''), role (``tourist''), or profession (``lawyer''). A larger vocabulary means more naming choices, which then results in higher variation across subjects. The mirror argument applies to less familiar kinds of objects such as animals.

We find evidence for both hypotheses in our analysis of the ManyNames ZH data, and suggest how to reconcile the two.

\section{Background} \label{sec:background}

\paragraph{Object naming in Psycholinguistics and Cognitive Science.}
Naming an object involves the selection of a specific term to refer to it \citep{silberer2020object}. In our daily life, it's common for objects to simultaneously fit into several categories; for instance, a given baby can belong to multiple overlapping categories like PERSON, FEMALE, BABY, and GIRL, among others. The names associated to these categories (e.g.\ ``human'', ``person'', etc.) are then all valid alternative names for this baby \cite{brown1958shall}, resulting in variation.
By far the most examined dimension of variation has been the taxonomic one, starting with seminal work by Rosch and colleagues \cite{rosch1976basic}. This line of work divides categories into three levels: superordinate (e.g., ANIMAL), basic (e.g., DOG), and subordinate (e.g., ROTTWEILER). Rosch and subsequent work showed that, in general, people prefer names corresponding to the basic level, which is hypothesized to represent a good balance between the specificity and distinctiveness of the categories \citep{murphy1985category}. 
However, another very prominent source of variation is so-called cross-classification \cite{Ross1999,shafto2011probabilistic}, whereby objects belong to different categories that are not hierarchically organized but merely overlap (for instance, WOMAN and TOURIST). 

In Cognitive Science, picture naming is the most widely used experimental paradigm for aspects related to naming \citep{snodgrass1980standardized,Brodeur2010,liu2011timed,Alario1999,tsaparinaRussianNormsName2011}.
Participants are presented with a visual stimulus and asked to produce the first name that comes to mind.
The resulting datasets are called picture-naming norms, or naming norms for short.
An important point for our purposes is the fact that, typically, due to the research goals of most of this research, the stimuli are prototypical pictures that represent categories, rather than the varied kinds of instances that one encounters in real life. Therefore, subjects reach a very high agreement in this task in terms of lexical choices \citep{rossion2004revisiting}. 
This is also true for the few naming norms that exist for Mandarin Chinese \cite{liu2011timed,weekes2007predictors,zhou2017color}.
ManyNames \citep{silberer2020object,silberer2020humans} draws inspiration from this paradigm but uses real-world images that show objects in their natural contexts, which elicits much more variation.

Previous work has shown that properties related to lexical access (word frequency, age of acquisition) affect the production probability of names \cite{Alario1999,Brodeur2010,snodgrass1980standardized,tsaparinaRussianNormsName2011}: All else being equal, more frequent words and words acquired earlier are preferred.
Although less studied, research also shows that the properties of the pictured objects influence people's naming choices; objects that are less typical for the category denoted by the most produced name trigger higher variation \cite{snodgrass1980standardized,gualdoni2022woman}.
People's naming choices are more varied for objects that are less typical for a frequent name. We focus on a different factor, namely familiarity (see below for more information). 

\paragraph{Object naming in Computer Vision and Language \& Vision.} 
The task of Object Recognition in the realm of Computer Vision aims to identify and classify objects, assigning them a single ground-truth label from a pre-defined vocabulary \cite{everingham2015pascal,russakovsky2015imagenet,kuznetsova2020open}.
While this approach resembles picture naming, most of this research overlooks linguistic aspects related to natural language, in particular the fact that categories overlap and that different words can be used for a single category. 
The ManyNames dataset, from which we draw our images, was built a.o.\ as a response to this issue \cite{silberer2020humans}. 

Several resources in Language \& Vision (a field at the intersection between Computer Vision and Computational Linguistics) have collected referring expressions for real-world images. While existing resources like RefCOCO and RefCOCO+ \citep{yu2016modeling}, Flickr30K-Entities \citep{plummer2015flickr30k}, and VisualGenome \citep{krishna2017visual} can be a source naming data for objects in context, they lack sufficient data for a systematic assessment of the variability and stability of object naming. 
In contrast, ManyNames focuses on object names in isolation and elicits many more names for the same object from different subjects than any other resource to date. 

\begin{figure}[htb]
     \centering
     \includegraphics[scale=0.2]{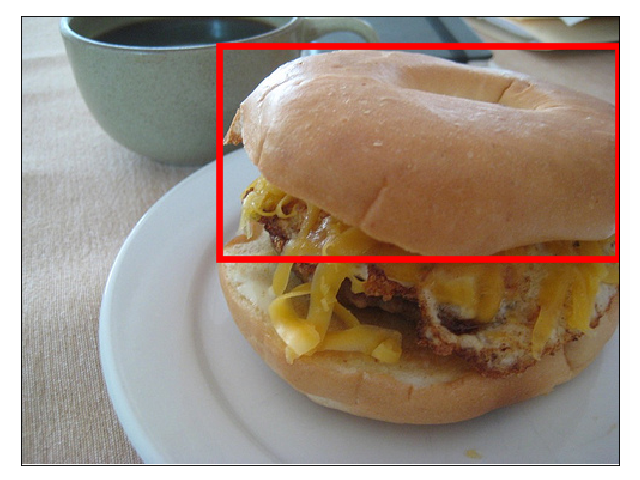}
     \caption{Image of a bagel.}
     \label{fig:bagel}
\end{figure}

\paragraph{Familiarity and naming behavior.} In psycholinguistic research, traditionally familiarity has been assessed through rating tasks, where participants assign ratings on a scale to indicate the degree of familiarity they have with the stimuli \citep{snodgrass1980standardized,sirois2006picture,boukadi2016norms}. 
Participants are instructed to consider objects encountered frequently in their daily lives as familiar, while categorizing rare or infrequently encountered objects as unfamiliar.
In picture naming norms, familiarity, along with factors such as name agreement, lexical frequency, imageability, age of acquisition, and visual complexity, has been identified as a predictor of naming latencies%
\footnote{The time it takes for a subject to start producing a name for a given stimulus.}
for both object and action pictures \citep{snodgrass1980standardized,sirois2006picture,liu2011timed}. 
It has also been shown to affect lexical choice \citep{anaki2009familiarity}. For example, when presented with an object like Figure \ref{fig:bagel}, individuals who describe it as  ``bread'' or ``burger'' likely possess limited prior knowledge about different types of bread in the USA. On the other hand, if someone readily identifies the object as a  ``bagel'', it suggests a higher level of familiarity.

Familiarity has also been related to vocabulary size for a given domain.
In a study by \citet{gatewood1984familiarity}, fifty-four American college students ranked their familiarity and knowledge about four semantic domains: musical instruments, fabrics, trees, and hand tools. They were asked to list all the categories of each domain they could think of in a free-recall task. The results showed that familiarity strongly predicts the size of salient vocabulary in each domain. 

The relationship between familiarity and naming variation, specifically, remains an open question, as results have varied across multiple studies.
A large study of picture-naming norms \citep{krautz2022linguapix} found that naming agreement and accuracy were higher for those images that participants were familiar with. The same was found Tunisian Arabic data in \citet{boukadi2016norms}, and for Mandarin Chinese in \cite{liu2011timed, zhou2017color}. 
However, a study of picture-naming norms for Canadian French by \citet{sirois2006picture} revealed no relationship between naming agreement and object familiarity.
Furthermore, note that familiarity has been shown to be culturally specific and may vary across different language communities \citep{boukadi2016norms}. For instance, the Mexican dish guacamole may not be familiar within Chinese-speaking contexts.

In our study, we focus on the level of familiarity among Mandarin speakers regarding the objects sampled from the ManyNames dataset, and how this factor influences their naming variation. The stimuli thus are very different from the ones traditionally used in psycholinguistics, and can shed complementary light on the relationship between familiarity and naming variation. 
We also experiment with a corpus-derived measure of familiarity instead of using human ratings.

\section{The ManyNames ZH dataset}
 

\subsection{Source dataset: ManyNames} \label{sec:manynames}

Our ManyNames ZH dataset is based on the verified ManyNames dataset (ManyNames v2).%
\footnote{Available at \url{https://github.com/amore-upf/manynames}.}
The original ManyNames dataset \citep{silberer2020object} provides 36 crowd-sourced annotations for 25K object instances obtained from VisualGenome \citep{krishna2017visual}. The objects are categorized into seven domains: ANIMALS\_PLANTS, BUILDINGS, CLOTHING, FOOD, HOME, PEOPLE, and VEHICLES. 
The annotations were obtained through an elicitation task conducted on Amazon Mechanical Turk (AMT), where participants were instructed to produce the first name that came to mind describing the object outlined by the red bounding box.
To address the presence of noise in the data, a second version of ManyNames was created \citep{silberer2020humans}. 
Specifically, another round of annotation tasks was conducted on AMT to clean naming errors. 
Analysis revealed that most inadequacies correspond to referential issues (e.g., subjects responding ``ball'' for the image in Figure \ref{fig:polar bear}; in Mandarin Chinese, no subject produced ``ball'', but instead they produced ``acrobatics'').
We used the English annotations to select a balanced sample of stimuli, as explained next.

\subsection{Image sampling} \label{sec: sampling}

\begin{figure}[hbt]
\centering
\begin{tikzpicture}[node distance=2.2cm, every node/.style={rectangle, draw, align=center, font=\small, text width=1.5cm, minimum width=1.7cm}]

\node (start) {Clear object};
\node [right of=start] (process1) {Race and ethnicity variation};
\node [right of=process1] (end) {Automatic sampling};

\draw[->] (start) -- (process1);
\draw[->] (process1) -- (end);

\end{tikzpicture}
\caption{Image sampling procedure.}
\label{fig: sampling_procedure}
\end{figure}
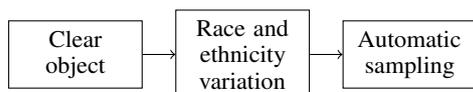

ManyNames consists of 1319 images, sampled in 3 steps illustrated in Figure~\ref{fig: sampling_procedure}.

In Step 1, we filtered unclear images from ManyNames v2 to mitigate referential issues, keeping only images where at least 75\% out of the subjects agree on the object being targeted. 

In Step 2, we made an intervention in the PEOPLE domain to ensure variability in race and ethnicity within the selected images. The ManyNames dataset primarily represents Western culture, particularly American culture, so a simple random choice would produce mostly images of white people. 
We used Computer Vision models to determine the race of individuals in the images, in particular the OpenCV \citep{opencv_library} and Deepface \citep{serengil2020lightface} libraries. 
Given noise in the automatically identified images, two authors of the paper annotated the identified images of non-white people.%
\footnote{The tools we use are trained with images in facial datasets (e.g., see \citealt{taigman2014deepface}). Generally, efforts are made to include clear and well-captured face images in these datasets. The human faces in our images are not always distinctly presented or complete, posing challenges for automatic identification using Computer Vision tools.}
A third author resolved discrepancies (see details in Appendix \ref{sec:app-sampling}). 
Images identified as picturing Middle-Eastern, Latino Hispanic and Indian people resulted in low inter-annotator agreement. 
We therefore included only images of Black and Asian individuals. We further randomly sampled an equal number of images depicting white people, paired on the basis of sharing the same top name (name most frequently produced by the subjects in ManyNames; for instance, it was ``woman'' for the image in Figure \ref{fig:woman}) and falling within the same variation band (see Step 3; also see Table \ref{tab: race balance} in Appendix \ref{sec:app-sampling} for statistics of the images). 
In total, we sampled 186 images in this step, with 93 non-white and 93 white individuals.

Most images in ManyNames have low variation; there is a prevalence of top names with mid-lexical frequency; and an imbalanced distribution across domains, with the majority of images belonging to the HOME domain (see Table \ref{tab: number of images by domain comparison} in Appendix \ref{sec: image sampling statistics}). 
Step 3 consisted in applying a sampling procedure to obtained a more balanced representation of naming variation, lexical frequency, and domains (details in Appendix \ref{sec:app-sampling}).%
\footnote{We also noticed that there was an image with the topname ``shoe'' in the PEOPLE domain, and removed it.}

\subsection{Data collection}
The collection of object names was obtained via crowdsourcing tasks on both Prolific\footnote{\url{https://www.prolific.co/}.} and AMT\footnote{\url{https://www.mturk.com/}.}.
The 1319 images were randomly divided into 7 lists, with participants being assigned randomly to one of the 7 lists. On average, it took approximately 40 minutes for a participant to complete the entire experiment.%
\footnote{In addition to collecting free names, there was a second part of the experiment that collected names after seeing a classifier. This second set of data was for a different study.}
The experiment interface and the instructions for annotators are included in Appendix \ref{sec: experiment procedure}. 

We also collected demographic data about the participants (detailed information in Appendix \ref{sec: demographic data}).
They were 146 Mandarin Chinese native speakers (61 females,
82 males, 1 non-binary individual and 2 participants with unknown gender). They ranged in
age from 18 to 50 years old, with 70\% belonging to the 18-35 age group.

We experienced difficulties obtaining data from Chinese speakers from these platforms because they prevail in Europe and USA, but not in China. On Prolific, a small portion of participants answered the questions in Cantonese or even English. On AMT, when we filtered for Mandarin Chinese, very few participants could see the task, so we had to remove the filter, resulting in most responses being in English. In the end, we collected data from 370 participants on AMT but could keep only 17. This is an example of the difficulties involved in building datasets for languages other than English. 

\subsection{Post-processing}\label{sec: post processing}

We post-processed the data to remove noise. 
First, we removed incorrect responses according to the criteria used in ManyNames. 
The four primary types of inadequate annotations are: referential (``named object not tightly in a bounding box''), visual recognition (``named object mistaken for something else it’s not, as in bear-dog''), linguistic (such as ``dear'' for ``deer'') and others \cite{silberer2020humans}. 
We used Google Translate to convert the identified mistaken English names in ManyNames v2 to Mandarin and excluded matching responses from the Chinese data. 

\begin{CJK}{UTF8}{gbsn}
Second, we converted responses in Pinyin, the primary romanization system for Standard Mandarin Chinese, into corresponding Chinese characters. 
We also eliminated responses containing expressions for uncertainty e.g., ``不知道'' (``I don't know''), and removed punctuation and non-Mandarin words.

Third, we used spaCy POS (part-of-speech) tagging \cite{honnibal2017spacy} to identify and remove adjectives in the responses, resulting in responses containing head words only, such as ``狗''(dog) instead of ``黑狗''(black dog) and ``小狗''(little dog).

Lastly, in the CLOTHING domain, despite the post-processing in Step 1, we still noticed errors related to subjects referring to the wearers rather than the clothing item. 
This is a common issue; \citet{silberer2020humans} hypothesize that it is due to people being much more salient than clothes for humans. 
We created a list of names for the PEOPLE domain by collating all the responses, manually excluded those associated with clothing, and filtered responses in the CLOTHING domain according to the cleaned list.
Note that despite this procedure some noise in the data remains, such as the name “杂技” (``acrobatics'') for the image in Figure \ref{fig:polar bear}. 
\end{CJK}

\begin{table*}[t]
\centering
\begin{tabular}{lccccccc}
\hline
Domain & N±std & H±std & F±std & \#Img & Voc.\ Size & Comp.\ Voc.\ Size \\
\hline
buildings & 8.0±3.1 & 2.3±0.9 & 2.9±0.5 & 170 & 503 & 423 \\
people & 7.2±2.1 & 2.2±0.5 & 3.9±0.4 & 320 & 501 & 284 \\
clothing & 6.8±2.1 & 2.2±0.6 & 2.9±0.3 & 145 & 295 & 281 \\
food & 6.2±2.4 & 1.9±0.8 & 2.8±0.3 & 136 & 269 & 269 \\
home & 6.0±3.0 & 1.7±0.9 & 2.9±0.4 & 203 & 556 & 414 \\
vehicles & 5.4±2.7 & 1.6±0.8 & 3.3±0.5 & 191 & 334 & 259 \\
animals\_plants & 4.1±2.2 & 1.2±0.7 & 3.1±0.5 & 154 & 212 & 192 \\
\hline
all & 6.4±2.8 & 1.9±0.8 & 3.2±0.6 & 1319 & 2670 & 2122 \\
\hline
\end{tabular}
\caption{Descriptive statistics for ManyNames ZH. Columns from left to right: domain, number N of distinct names per object (mean ± standard deviation); naming variation H (mean ± standard deviation)); familiarity F (mean ± standard deviation); total number of images (\#Img); vocabulary size (total name types); comparable vocabulary size (total name types calculated by randomly subsampling 136 images from all domains).}
\label{tab: descriptive stats}
\end{table*}

\subsection{Results}
Table~\ref{tab: descriptive stats} presents descriptive statistics for the entire dataset as well as for each of the seven domains (see next section for how naming variation and familiarity were computed). 
There are clear differences in terms of naming variation across domains, with BUILDINGS, PEOPLE and CLOTHING having higher naming variation than FOOD, HOME, VEHICLES and especially ANIMALS\_PLANTS. Instead, mean familiarity is similar across domains except for PEOPLE, with 3.9 compared to around 3.1 in other domains. 
The last column in Table~\ref{tab: descriptive stats} contains the comparable vocabulary size, obtained by randomly downsizing all domains to the smallest domain (sampling 136 images for all domains). Vocabulary size is largest in BUILDINGS and HOME; ANIMAL\_PLANTS has the lowest vocabulary size.%
\footnote{HOME is a heterogeneous domain, so it is expected to have a large vocabulary size. We instead have no explanation for the large vocabulary size in the BUILDINGS domain at present. Also note that, even though the domain is called ANIMAL\_PLANTS, the vast majority of the images in that domain correspond to animals.}

\section{Analysis}
\label{sec:analysis-method}

\paragraph{Estimates for variation and familiarity.} As standard in picture norms, naming variation for objects was estimated in terms of the entropy H of the responses. \citet{snodgrass1980standardized} introduced this metric and defined as in Eq.~\ref{formula: entropy}, where k refers to the number of different names given to each object and $p_i$ is the proportion of annotators giving each name. 

\begin{align}
    H = \sum_{i=1}^{k} p_i \log_{2}\left(\frac{1}{p_i}\right) \label{formula: entropy}
\end{align}

In this study, we use lexical frequency as a proxy for familiarity, based on the established positive relationship between familiarity and frequency \citep{boukadi2016norms, tanaka2011word}.
We aim at modeling the familiarity of kinds of objects represented in the images. As mentioned in Section \ref{sec:background}, in naming norms typically the objects are highly prototypical of a single named category. Instead, our stimuli are real-world images that are not always prototypical for a single salient category.
We use the naming responses as proxies for the categories that a given stimulus belongs to, and define familiarity as the weighted average of lexical frequency, as defined in Eq.~\ref{formula: familiarity}. 
Here $N$ is the set of responses for a given stimulus, $f(n)$ is the corpus-based frequency of name $n$, and the weighting factor $p(n)$ the proportion of subjects that produced that name. 
Frequency (in logarithm of base 10) for names was extracted from SUBTLEX-CH, a subtitle corpus of Mandarin Chinese \citep{cai2010subtlex}. 
For names not found in the corpus, we assign the average frequency of the remaining names associated with that object to them. 

\begin{align}
    F := \sum_{n \in N} f(n) \cdot p(n)
    \label{formula: familiarity}
\end{align}

\paragraph{Regression model.} We fitted a linear mixed-effects regression model with naming variation as the outcome variable and fixed effects for familiarity, domain, and their interactions.
All predictors were centered so that the reference level for each predictor is the overall mean across all levels of that predictor. 
The inclusion of the domain as a fixed effect allowed for the examination of potential systematic variations in naming across different domains.
The interaction between familiarity and domain was included to explore whether the relationship between naming variation and familiarity is domain-dependent.
The lists assigned to participants were treated as random intercepts.
All analyses were performed using Bayesian inference methods, using the brms-package \citep{burkner2021bayesian} of R (version 4.3.0, \citealt{r2021}).%
\footnote{Model in brms syntax: H $\sim$ familiarity * domain + (1$|$ list).}

\begin{table*}[htb]
\centering
\begin{tabular}{lrrc}
\hline
Variable & Estimate & Est.\ Error & 95\% CI \\
\hline
Intercept & 1.81 & 0.06 & [1.68, 1.94] \\
Familiarity & \textbf{-0.55} & 0.05 & [-0.65, -0.46] \\
Domain-animals\_plants & \textbf{-0.72} & 0.05 & [-0.83, -0.61] \\
Domain-home & \textbf{-0.38} & 0.06 & [-0.49, -0.27] \\
Domain-food & \textbf{-0.24} & 0.08 & [-0.40, -0.07] \\
Domain-vehicles & \textbf{-0.12} & 0.05 & [-0.22, -0.03] \\
Domain-buildings & \textbf{0.27} & 0.06 & [0.15, 0.39] \\
Domain-clothing & \textbf{0.42} & 0.07 & [0.28, 0.56] \\
Familiarity: home & \textbf{-0.44} & 0.11 & [-0.65, -0.24] \\
Familiarity: food & -0.20 & 0.17 & [-0.53, 0.13] \\
Familiarity: animals\_plants & -0.19 & 0.11 & [-0.40, 0.03] \\
Familiarity: buildings & 0.01 & 0.11 & [-0.21, 0.23] \\
Familiarity: vehicles & 0.19 & 0.09 & [-0.00, 0.36] \\
Familiarity: clothing & \textbf{0.55} & 0.15 & [0.26, 0.84] \\
\hline
\end{tabular}
\caption{Estimates of fixed effects when predicting naming variation (H) as a function of familiarity, domain, and the interaction between familiarity and domain. The last column shows the credible interval. Effects with CIs that do not straddle 0 are boldfaced.}
\label{tab:estimates}
\end{table*}

\section{Results}

Fixed effect estimates are shown in Table~\ref{tab:estimates}, where effects whose credible intervals (CI) do not cross 0 are boldfaced. 
The observed overall relationship between familiarity and naming variation aligns with H1: higher familiarity with a particular kind of object is associated with lower naming variation.

\begin{figure}[htbp]
  \centering
  \includegraphics[width=0.5\textwidth]{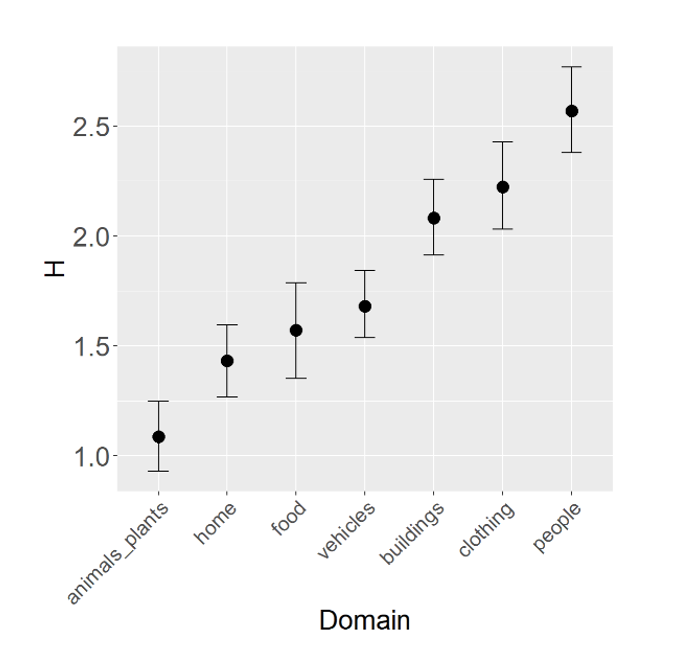} 
  \captionsetup{justification=centering, position=below}
  \caption{Predicted H of the domains covered in ManyNames ZH.}
  \label{fig:predicted H of seven domains}
\end{figure}

However, the model also suggests that variation is very different across domains. The domains, arranged in ascending order of naming variation, are as follows: ANIMALS\_PLANTS, HOME, FOOD, VEHICLES, BUILDINGS, CLOTHING, and PEOPLE (see Figure \ref{fig:predicted H of seven domains} for a visualization of model predictions for domains). Recall from Table \ref{tab: descriptive stats} that PEOPLE has the highest mean familiarity, and it also exhibits the highest model-predicted variation when holding other factors constant; and the converse for ANIMAL\_PLANTS. This supports H2: for domains that we are highly familiar with, we develop a larger vocabulary, and more lexical choices result in higher variation.

Furthermore, when examining the relationship between naming variation and familiarity across domains, we observe that CLOTHING is the only domain in which a higher familiarity of an object tends to increase, rather than decrease, naming variation. 

\begin{figure*}[htb]
  \centering
  \begin{minipage}[b]{0.45\linewidth}
    \centering
    \includegraphics[width=\textwidth]{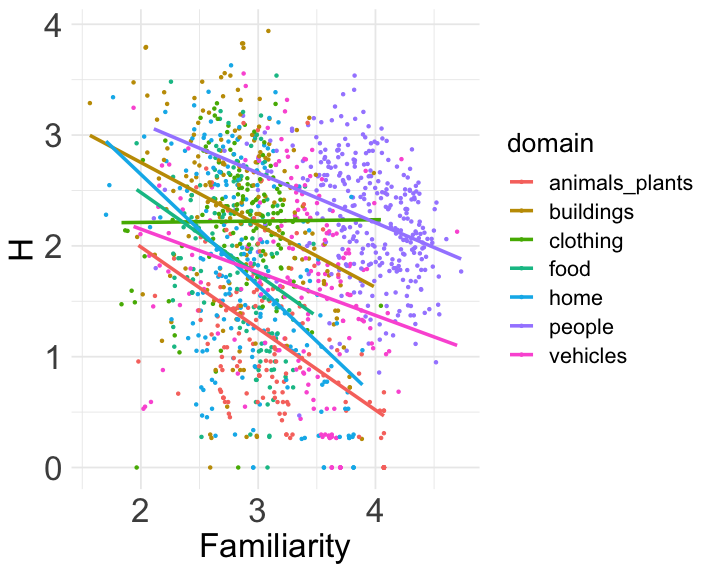}
    \caption{Effect by domain with a linear model.}
    \label{fig: fitted lines}
  \end{minipage}
  \hspace{0.05\linewidth}
  \begin{minipage}[b]{0.45\linewidth}
    \centering
    \includegraphics[width=\textwidth]{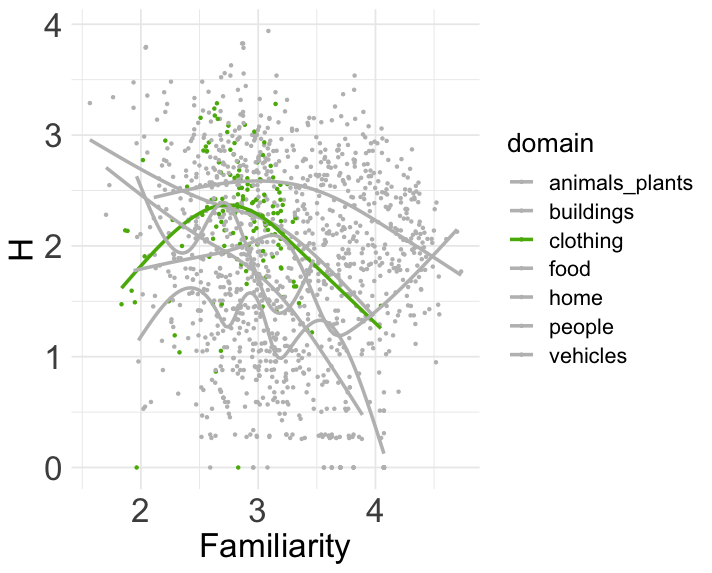}
    \caption{Effect by domain using a GAM.}
    \label{fig: fitted curve}
  \end{minipage}
\end{figure*}

\section{Discussion}
\begin{CJK}{UTF8}{gbsn}
Our results suggest that, in general, higher familiarity predicts lower naming variation (Hypothesis 1) when Mandarin Chinese speakers name visually presented objects. This indicates that people tend to converge on a common name for kinds of objects they're more familiar with.
For instance, in the ANIMALS\_PLANTS domain, people exhibit relatively low naming variation when referring to dogs (see Figure \ref{fig:dog}, where ``dog'' was produced by 21 out of 23 subjects).
We hypothesize that this can be attributed to the prevalence of dogs as pets in our daily lives.
Instead, we are less familiar with e.g.\ bears; in Figure \ref{fig:polar bear}, people use "北极熊" (``polar bear'') and “熊” (``bear'') in almost equal proportion, and they also use the more general term “动物”(``animal''). Note that some people do not correctly identify the kind of animal, naming it instead “狗” (``dog'') or “海马” (``seahorse'').%
\footnote{\citet{silberer2020humans} noted that subjects preferred the basic level term even if they risk being wrong (e.g.\, in cases where the gender of the person was not clear some subjects produced ``man'' or ``woman'' as opposed to ``person'').}

However, an intriguing contradiction to this finding emerges when we consider the effect of different domains on naming variation. Although humans are arguably more familiar with people than with animals (conjecture supported by the data in Table \ref{tab: descriptive stats}), naming variation within the PEOPLE domain is actually much higher than that within the ANIMALS\_PLANTS domain.%
\footnote{\citet{silberer2020object} found the same for English.}
At the domain level, thus, naming variation actually \textit{increases} with familiarity, in accordance with Hypothesis 2 and against Hypothesis 1. 
This is consistent with \citet{gatewood1984familiarity}, which as discussed in Section \ref{sec:background} found salient vocabulary size to be positively correlated with familiarity in American English, for domains such as musical instruments. 
Chinese similarly seems to have a richer vocabulary for people as opposed to e.g.\ animals (see Table \ref{tab: descriptive stats}). 
This effect can be due to the fact that when we interact a lot with a given category of objects, like that of people, we need to develop a richer vocabulary to draw finer-grained distinctions within the category and facilitate communication.
A larger vocabulary affords more opportunities for naming variation to arise. 

Additionally, we also find evidence of the two factors being at play within the CLOTHING domain. While a linear regression model suggests that naming variation increases or plateaus in the CLOTHING domain (see Figure \ref{fig: fitted lines}), fitting the data to a generalized additive model uncovers a clear convex curve (see Figure \ref{fig: fitted curve}).%
\footnote{The figure exhibits a smooth curve fitted to a scatter plot using geom\_smooth() in ggplot2 \citep{ggplot2} with the method = ``gam'' argument and formula H $\sim$ s(familiarity, by = domain).}
Manual inspection revealed that in the low-variation, low-familiarity area we have specific but unfamiliar objects like bowties; in the low-variation, high-familiarity area there are specific and familiar objects like t-shirts; and in the high-variation, mid-familiarity area there are types of clothes that are neither unfamiliar nor very familiar for Chinese speakers, like the jackets of masculine Western suits, which receive names such as “套装” and “西装” (``suit''), “衣服” (``clothes''), “外套” (``jacket''), or “西服” (``Western clothes''). 

We thus find evidence for both hypotheses, which however play at different levels of granularity. At the level of a specific object, higher familiarity with that object's category implies lower variation because people converge on the same label for the object. At the level of the domain or supra-category, instead, higher familiarity implies higher variation because of the richer vocabulary available for speakers. 
\end{CJK}
\section{Conclusion}

In this paper, we have introduced ManyNames ZH, a new Language and Vision dataset designed for the task of Object Naming in Mandarin Chinese. The new dataset is the result of crowdsourcing names in Mandarin Chinese, based on the images from the English ManyNames dataset, with pre- and post-processing steps. ManyNames ZH consists of a carefully curated subset of 1319 images, each accompanied by an average 20 names provided by different human annotators. It allows the community to expand the empirical basis of findings on naming, by including a major language from a typologically different family than English. 
With the availability of ManyNames subsets in three languages, English, Catalan \cite{orfila2022cat}, and Mandarin Chinese, researchers can also conduct cross-linguistic studies and comparative analyses on object naming. 

With this new dataset, we have explored the relationship between object familiarity and the degree of naming variation.
We observe two opposite factors at play. On the one hand, when familiarity with objects in a given supra-category or domain increases (such as with the PEOPLE domain), vocabulary size correspondingly increases, too. This affords higher naming variation because it gives speakers more options to choose from. On the other hand, within a given category, more familiar sub-categories will afford conventionalization of the label used to talk about it, which elicits lower naming variation. 
This helps explain conflicting results found in Psycholinguistic studies on naming, which found the effect of domain on vocabulary size \citep{gatewood1984familiarity}; a negative correlation between familiarity and variation variation \citep{krautz2022linguapix,boukadi2016norms}; and no relation between the two factors  \citep{sirois2006picture}, respectively. 

Our analysis is based on a snapshot of Mandarin Chinese in which the vocabulary is frozen and we only observe the use. However, the patterns observed result from the dynamic evolution of vocabulary over time. 
Our results suggest that the need to frequently talk about a given kind of object triggers the development of a richer vocabulary that accounts for relevant distinctions within that broad class; and that higher communication about a specific kind of object triggers the convergence on a single label. Future work should test this hypothesis empirically. 


\section*{Limitations}

Our dataset still contains noise despite the post-processing efforts, particularly in the PEOPLE and CLOTHING domains. Challenges arise from referential errors, as well as the inclusion of non-noun words in the dataset. Additional steps, such as further semi-automatic or crowdsourcing-based filtering (as was done for the English ManyNames) could help address these issues. 

Also, given the limited availability of native Mandarin Chinese speakers on the platforms we utilized, we were only able to gather an average of 20 annotations per image. In comparison, the English ManyNames dataset contains an average of 31 annotations per image. 
As mentioned above, this showcases the difficulties of building resources for non-Western languages. 

It is also important to note that the images from the original ManyNames dataset primarily reflect the cultural background of the USA. 
We made an effort to balance racial representation in the PEOPLE domain, but we did not address cultural biases in other domains that are also heavily culture-dependent, in particular FOOD and CLOTHING, as we deemed it more difficult to do this with automatic means. 
Future work in Language and Vision needs to address cultural biases \cite{liu-etal-2021-visually}.

Finally, in our study, we used the weighted average of the lexical frequency of the responses as a measure of familiarity for objects. Alternatively, subjective ratings of familiarity by human participants can provide valuable insights and should be considered in future research. Also, there are individual differences in familiarity, and we provide a measure of overall expected familiarity within a culture, without taking into account these individual differences. We leave it to future work to investigate the relationship between familiarity and naming behavior at the individual level. 

\section*{Ethics Statement}
This paper complies with the \href{https://www.aclweb.org/portal/content/acl-code-ethics}{ACL Ethics Policy}. 
Quoting from the ACM Code of Ethics, we :(1) ``contribute to society and to human well-being, acknowledging that all people are stakeholders in computing'', by investigating how computational models can contribute to answer questions about how language works; (2) ``avoid harm'' by broadening the empirical basis of work on Language and Vision, introducing a new dataset for Mandarin Chinese; (3) are ``honest and trustworthy'' about our results and limitations; (4) ``attempt to be fair and take action not to discriminate'' by including considerations of race variability in our image sampling method (although future work should do more in including other sources of cultural variation); (5) ``respect the work required to produce new ideas, inventions, creative works, and computing artifacts'' by citing the related work that contributed to our work to the best of our knowledge; (6) ``respect privacy'' and (7) ``honor confidentiality'' by anonymizing the dataset prior to its public distribution. Like any work in AI and indeed in science and technology, of course, the results of our work can be used both for good and for bad.

\section*{Acknowledgements}
This project has received funding from the Ministerio de Ciencia e Innovación and
the Agencia Estatal de Investigación (Spain; ref. PID2020-112602GBI00/MICIN/AEI/10.13039/
501100011033).
We also thank the financial support from the Catalan government (SGR 2021 00470) and the Department of Translation and Language Sciences at Universitat Pompeu Fabra.


\bibliography{anthology,custom}
\bibliographystyle{acl_natbib}
\clearpage

\section*{Appendices}
\appendix
\section{Image Sampling Statistics}
\label{sec: image sampling statistics}

\begin{table}[htb]
\centering
\resizebox{0.9\textwidth}{!}{
\begin{tabular}{|c|c|}
\hline
ManyNames v2 & Sample  \\
\hline
\includegraphics[width=0.3\textwidth]{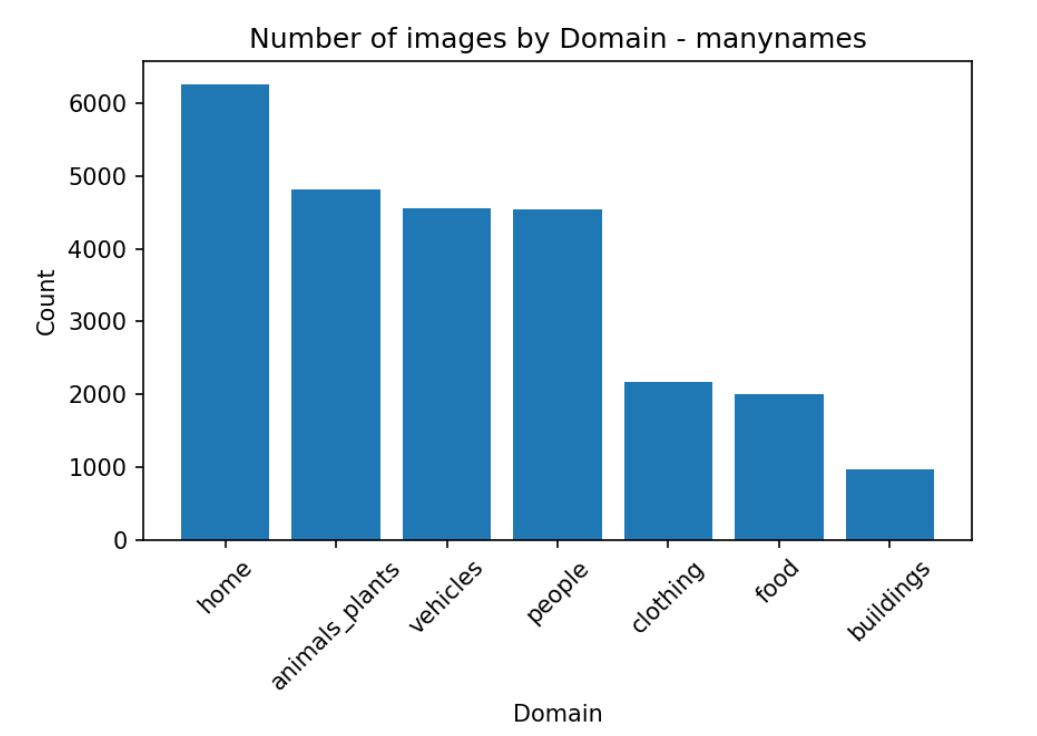} & \includegraphics[width=0.3\textwidth]{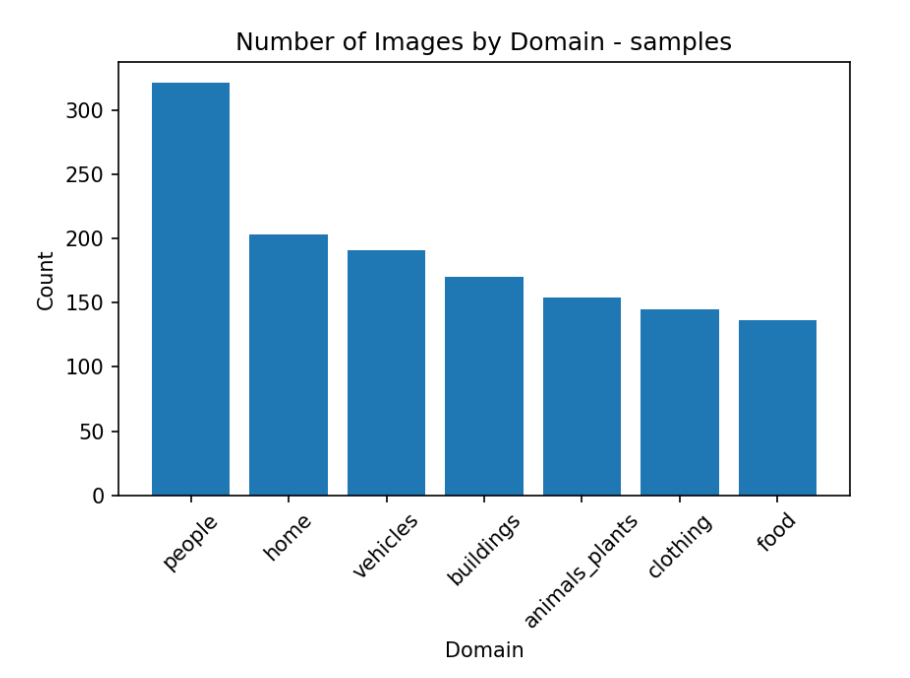} \\
\hline
\end{tabular}
}
\captionsetup{width=1.1\textwidth}
\captionsetup{singlelinecheck=false, justification=raggedleft}
\caption{Distribution of images across domains in ManyNames v2 and sample.}
\label{tab: number of images by domain comparison}
\end{table}

\begin{table}[htp]
\centering
\resizebox{0.9\textwidth}{!}{
\begin{tabular}{|c|c|}
\hline
ManyNames v2 & ManyNames ZH  \\
\hline
\includegraphics[width=0.3\textwidth]{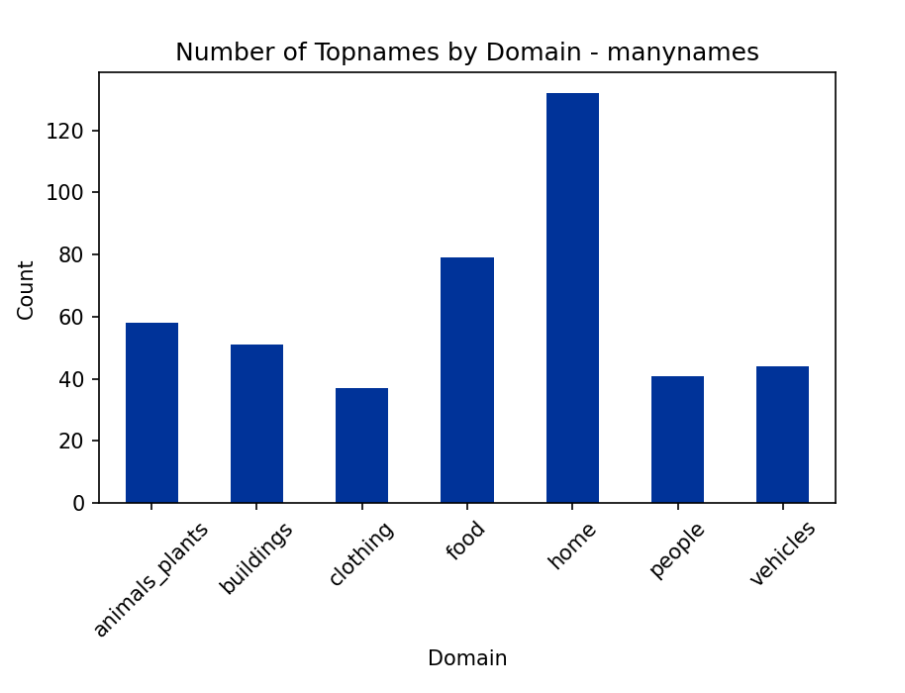} & \includegraphics[width=0.3\textwidth]{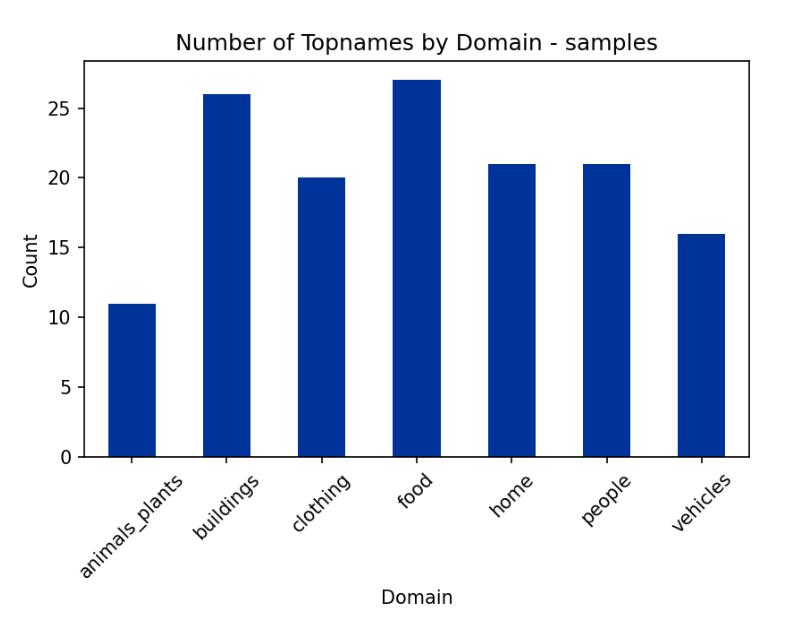} \\
\hline
\end{tabular}
}
\captionsetup{width=1.2\textwidth}
\captionsetup{singlelinecheck=false, justification=raggedleft}
\caption{Distribution of topnames across domains in ManyNames v2 and ManyNames ZH.}
\label{tab: number of topnames by domain comparison}
\end{table}

\begin{table*}[htb]
\centering
\resizebox{\textwidth}{!}{
\begin{tabular}{|b{0.11\textwidth}|c|c|c|}
\hline
Dataset & Corpus-based frequency & ManyNames-based frequency & Naming variation \\
\hline
ManyNames v2 & \includegraphics[width=0.25\textwidth,valign=c]{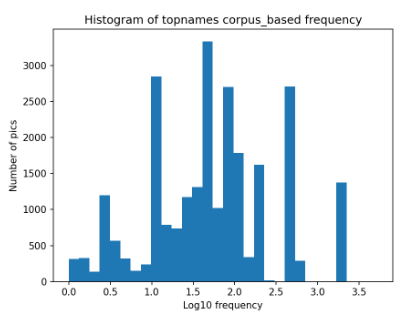} & \includegraphics[width=0.25\textwidth,valign=c]{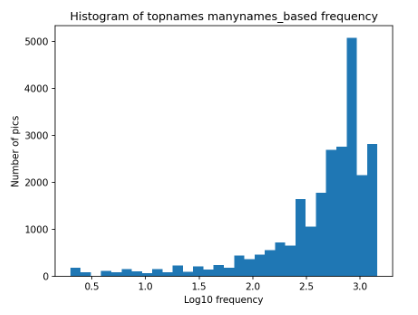} & \includegraphics[width=0.25\textwidth,valign=c]{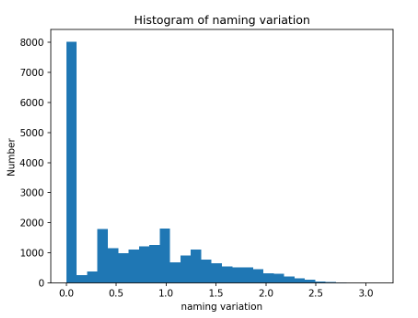}  \\
\hline
Sample & \includegraphics[width=0.25\textwidth,valign=c]{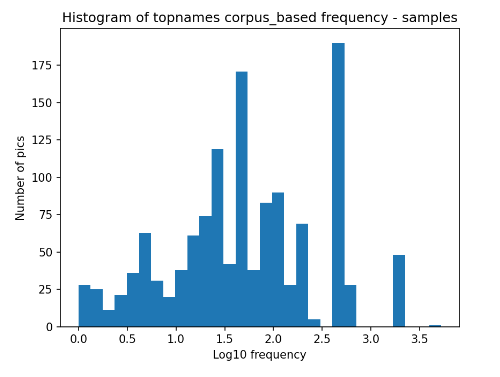} & \includegraphics[width=0.25\textwidth,valign=c]{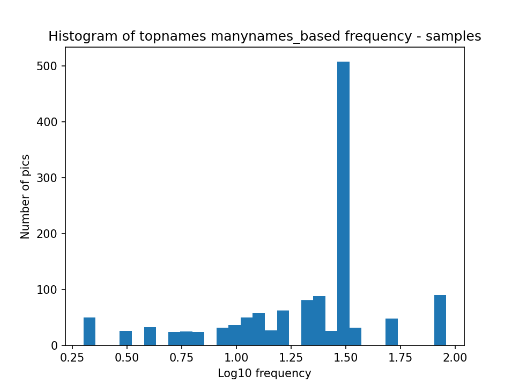} & \includegraphics[width=0.25\textwidth,valign=c]{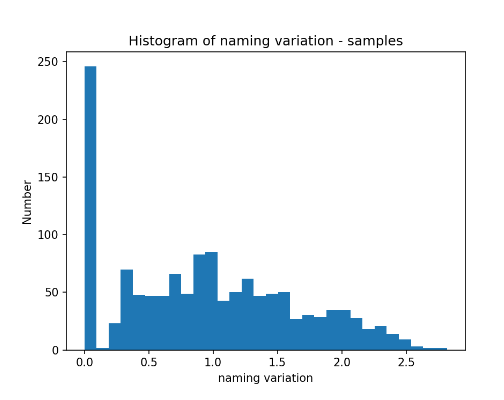}  \\
\hline
Sample- low frequency band & \includegraphics[width=0.25\textwidth,valign=c]{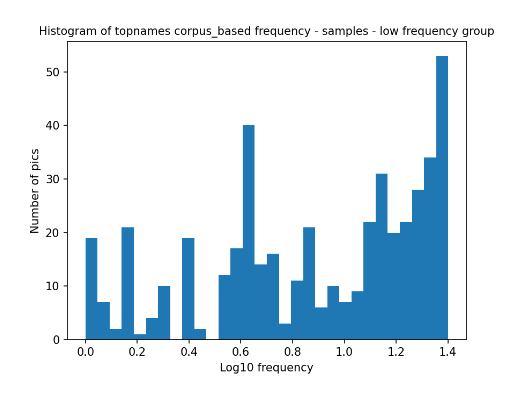} & \includegraphics[width=0.25\textwidth,valign=c]{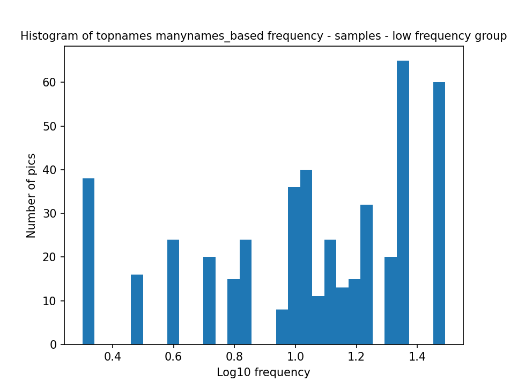} & \includegraphics[width=0.25\textwidth,valign=c]{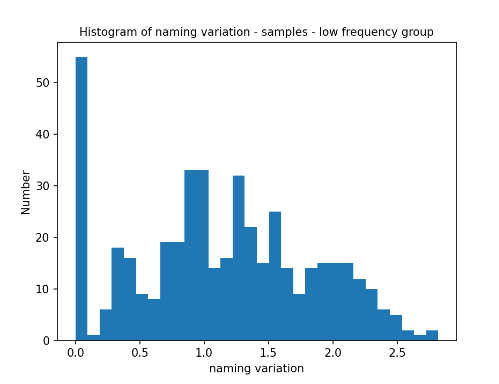} \\
\hline
Sample-mid frequency band & \includegraphics[width=0.25\textwidth,valign=c]{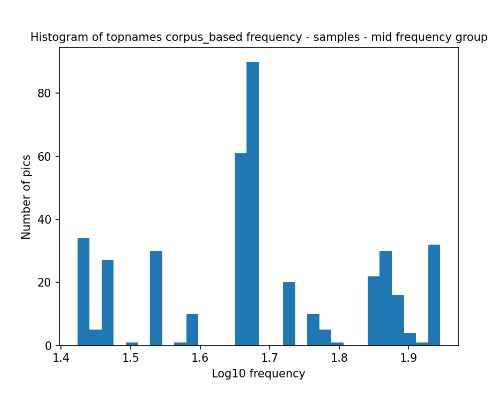} & \includegraphics[width=0.25\textwidth,valign=c]{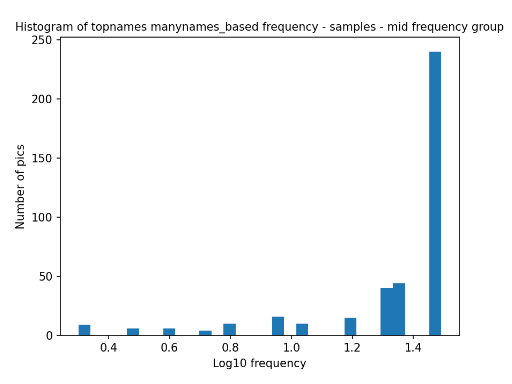} & \includegraphics[width=0.25\textwidth,valign=c]{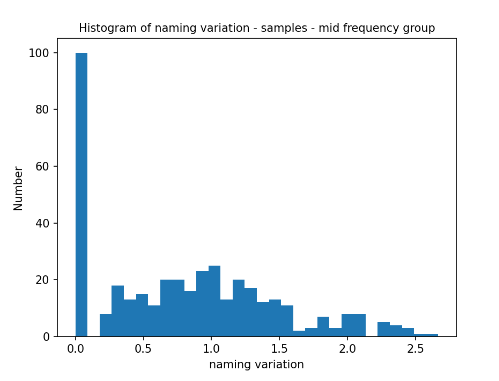}  \\
\hline
Sample-high frequency band & \includegraphics[width=0.25\textwidth,valign=c]{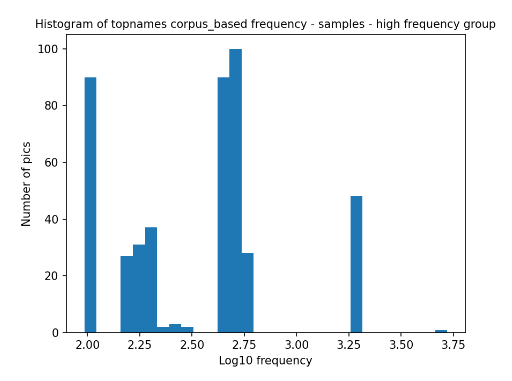} & \includegraphics[width=0.25\textwidth,valign=c]{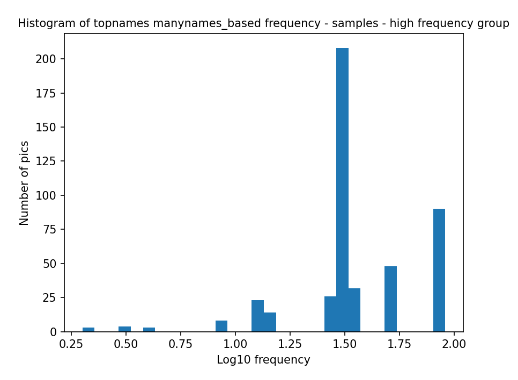} & \includegraphics[width=0.25\textwidth,valign=c]{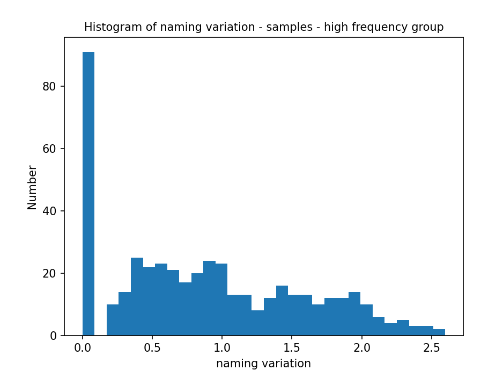}  \\
\hline
\end{tabular}
}

\caption{Distribution of ManyNames, sampled images and each frequency band of sampled images in terms of topname frequency (corpus-based) in logarithm of base 10, topname frequency (ManyNames-based) in logarithm of base 10, and naming variation.}
\label{tab:stats_image_sampling}
\end{table*}

\clearpage
\section{Details on sampling}
\label{sec:app-sampling}

Table \ref{tab: race balance} shows the distribution of non-white images. 

As for the automatic sampling, it consists of the following steps. First, we partitioned the images into three naming variation bands (low, mid, and high) using quantiles. Each band contained an equal proportion of the total images, resulting in approximately one-third of the images in each band.
Likewise, we divided the topnames into three frequency bands (low, mid, and high) based on their corpus-based frequency in the logarithm of base 10 using quantiles.
The frequency data were derived from SUBTLEX-US, a subtitle corpus of American English \cite{brysbaert2009moving}. Each frequency band also contained approximately one-third of the topnames. 

We initiated the image sampling from a specific domain (e.g., FOOD). 
Within the chosen domain, we focused on a particular frequency band (e.g., low frequency band).
Next, we randomly selected a single topname (e.g., ``cupcake'') from the selected frequency band.
For the chosen topname, we proceeded to sample 10 images from each of the low, mid, and high variation bands. 
If a variation band had fewer than 10 available images, we settled with all available ones and moved to the next variation band. 
We repeated this process of topname sampling until approximately 60 images were obtained for the selected frequency band. 
Following this, we repeated the sampling procedure for each frequency band within the selected domain, resulting in approximately 180 images obtained for each domain.
This entire procedure was then replicated for the remaining six domains.
Note that for the PEOPLE domain, we excluded previously sampled topnames from Step 2 to avoid duplication in this step (i.e., ``woman'', ``man'', ``girl'', ``boy'', ``child'' and ``skier'' in Table \ref{tab: race balance}). We then sampled additional images until reaching 10 images or the maximum available per variation band. However, if the number of images for a specific topname already exceeded 10 in Step 2, we did not sample any additional images for that topname.

\begin{table}[htb]
\begin{tabular}{|l|p{0.2\columnwidth}|p{0.2\columnwidth}|p{0.25\columnwidth}|}
\hline
Race & Low & Mid & High \\
\hline
Asian & 4 (``woman'': 3, ``man'': 1) & 38 (``woman'': 27, ``man'': 9, ``girl'': 2) & 39 (``woman'': 9, ``girl'': 9, ``boy'': 9, ``man'': 6, ``child'': 5, ``skier'': 1) \\
\hline
Black & 0 & 6 (``man'': 4, ``woman'': 2) & 6 (``boy'': 2, ``child'': 2, ``woman'': 2) \\
\hline
Total & 4 & 44 & 45 \\
\hline
\end{tabular}
\caption{Distribution of non-white images sorted by naming variation band; number out of parentheses is the number of images, and number in parentheses indicates the number of images with the corresponding top name.}
\captionsetup{}
\label{tab: race balance}
\end{table}

\section{Demographics}
\label{sec: demographic data}
\paragraph{Demographic questionnaire}
\begin{CJK}{UTF8}{gbsn}
\section*{中文物体命名：背景调查表}

实验之前需要填写一份背景调查。相关信息严格保密的，不会以任何方式与您的姓名或身份相关联。请尽您所能回答问题。如果您对这份问卷有任何问题或疑虑，请在继续填写之前发送邮件到：\textit{[email address]}

注意：标有星号（*）的问题是必答题。回答后才能进入下一步，谢谢您的合作！

\begin{enumerate}
  \item 您的年龄？* \\
  （）18-25 \\
  （）26-35 \\
  （）36-45 \\
  （）46及以上
  
  \item 您的性别？* \\
  \underline{\hspace{4cm}}
  
  \item 您的学历（包括在读）？* \\
  （）"高中及以下" \\
  （）"大专" \\
  （）"本科" \\
  （）"硕士研究生" \\
  （）"博士研究生及以上"
  
  \item 普通话是你小时候学习的第一种语言吗？* \\
  （）是 \\
  （）否
  
  \item 在15岁之前，您是否都在中国居住？* \\
  （）是 \\
  （）否
  
  \item 您还会说其他语言吗？* \\
  （）是 \\
  （）否 \\
  如果是，请写出其他语言中最精通的语言和对该语言的熟练程度（熟练程度供参考：入门、基础、中级、高级、母语）： \\
  参考示例：英语，高级 \\
  \underline{\hspace{4cm}}
  
  \item 在6岁之前，除了普通话之外，家里是否还有其他语言？*（包括方言） \\
  （）是 \\
  （）否 \\
  如果是，家里说的是什么语言（或方言）： \\
  \underline{\hspace{4cm}}
  
  \item 您是否在非汉语国家学习或工作过？* \\
  （）是 \\
  （）否 \\
  如果是，请说明居住时间最长的一个国家和大致居住的时间： \\
  参考示例：西班牙，3年
\end{enumerate}

\end{CJK} 

\section*{Translation}
\section*{Object naming in Mandarin Chinese: background questionnaire}

A background survey needs to be completed prior to the experiment. The relevant information is strictly confidential and will not be associated with your name or identity in any way. Please answer the questions to the best of your ability. If you have any questions or concerns about this questionnaire, please send an email to \textit{[email address]}
before proceeding.

Note: Questions marked with an asterisk (*) are mandatory. Thank you for your cooperation!

1. How old are you? *(Required)
\begin{itemize}
  \item 18-25
  \item 26-35
  \item 36-45
  \item 46 and above
\end{itemize}

2. What is your gender? *(Required)
\begin{tabular}{@{}ll@{}}
  \underline{\hspace{3cm}}
\end{tabular}

3. Please indicate your education level (including current status)* (Required)
\begin{itemize}
  \item "High school or below".
  \item "Vocational college"
  \item "Bachelor's degree"
  \item "Master's degree"
  \item "Doctoral degree or above"
\end{itemize}

9. Was Mandarin Chinese the first language you learned as a child? *(Required)
\begin{itemize}
  \item Yes
  \item No
\end{itemize}

10. Did you live in China until you were 15 years old? *(Required)
\begin{itemize}
  \item Yes
  \item No
\end{itemize}

11. Do you speak any other languages? *(Required)
\begin{itemize}
  \item Yes
  \item No
\end{itemize}

If yes, please write the most proficient of the other languages and the level of proficiency in that language (proficiency level for reference: Beginner, Basic, Intermediate, Advanced, Native):
Reference Example: English, advanced
\begin{tabular}{@{}ll@{}}
  \underline{\hspace{5cm}}
\end{tabular}

12. Before the age of 6, were there any other languages spoken at home besides Mandarin (including dialects)? * (Required)
\begin{itemize}
  \item Yes
  \item No
\end{itemize}

If yes, what language (or dialect) was spoken at home: 
\begin{tabular}{@{}ll@{}}
  \underline{\hspace{5cm}}
\end{tabular}

13. Have you ever studied or worked in a non-Chinese speaking country? *(Required)
\begin{itemize}
  \item Yes
  \item No
\end{itemize}

If yes, please indicate the country where you have lived the longest and the approximate length of residence:
Reference example: Spain, 3 years
\begin{tabular}{@{}ll@{}}
  \underline{\hspace{5cm}}
\end{tabular}

\clearpage

\begin{table*}[ht]

\centering

\begin{tabular}{|p{5cm}|l|c|c|}
\hline
Variable & Category & Frequency & Percentage \\
\hline
Age & 18-25 & 44 & 30.1\% \\
 & 26-35 & 58 & 39.7\% \\
 & 36-45 & 31 & 21.2\% \\
 & 46-50 & 13 & 8.9\% \\
\hline
Gender & Female & 61 & 41.8\% \\
 & Male & 82 & 56.2\% \\
 & Non-binary & 1 & 0.7\% \\
 & Unknown & 2 & 1.4\% \\
\hline
Educational level & High school or below & 3 & 2.1\% \\
 & Vocational college & 8 & 5.5\% \\
 & Bachelor's degree & 58 & 39.7\% \\
 & Master's degree & 53 & 36.3\% \\
 & Doctoral degree or above & 24 & 16.4\% \\
\hline
Mandarin Chinese as first language learned? & Yes & 139 & 95.2\% \\
 & No & 7 & 4.8\% \\
\hline
Live in China until 15 years old? & Yes & 120 & 82.2\% \\
 & No & 26 & 17.8\% \\
\hline
Speak any other languages? & Yes & 143 & 98.0\% \\
 & No & 3 & 2.0\% \\
\hline
Before the age of 6, were there any other languages spoken at home besides Mandarin (including dialects)? & Yes & 70 & 48.0\% \\
 & No & 76 & 52.0\% \\
\hline
Have you ever studied or worked in a non-Chinese speaking country? & Yes & 131 & 89.7\% \\
 & No & 15 & 10.3\% \\
\hline
\multicolumn{4}{|l|}{n = 146} \\
\hline
\end{tabular}
\caption{Descriptive statistics on the demographics of the participants in ManyNames ZH.}
\label{tab: demographic data}
\end{table*}

\clearpage
\section{Experiment Procedure}
\label{sec: experiment procedure}

\begin{figure}[hbt]
\centering
\begin{tikzpicture}[node distance=2cm, every node/.style={rectangle, draw, align=center, font=\small, text width=1.5cm, minimum width=1.5cm}]

\node (start) {Consent form};
\node [right of=start] (process1) {Background questionnaire};
\node [right of=process1] (process2) {Object naming};
\node [right of=process2] (end) {Naming with classifiers};

\draw[->] (start) -- (process1);
\draw[->] (process1) -- (process2);
\draw[->] (process2) -- (end);

\end{tikzpicture}
\caption{Experiment design}
\label{fig: exp design}
\end{figure}
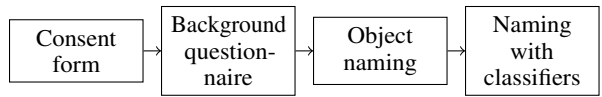

 Our experiment consisted of four sessions: consent form, background questionnaire, object naming, and object naming with classifiers. The last one, adapted from the third session, served for another study.
 
 Also, the initial pilot studies revealed that participants tended to use modifiers and numerical classifiers when describing objects. To address this, the instructions were modified to discourage the use of such linguistic elements. (see Appendix \ref{sec: experiment procedure} for experiment interface and instructions for annotators).

\begin{figure}[htbp]
  \centering
  \includegraphics[width=0.5\textwidth]{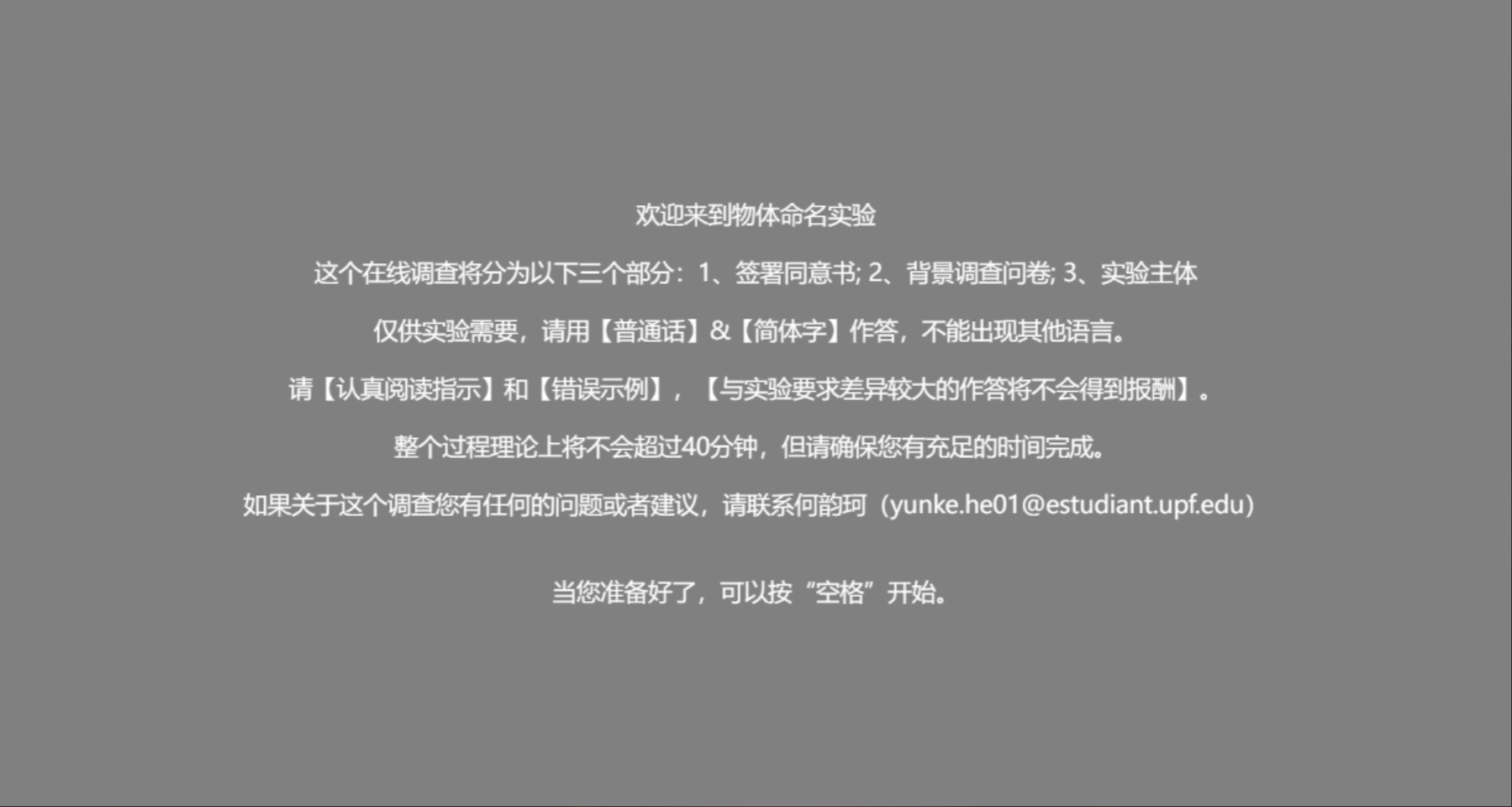} 
  \captionsetup{justification=centering, position=below}
  \caption{Introduction}
  \label{fig:introduction}
\end{figure}

\section*{Translation for Figure \ref{fig:introduction}}

Welcome to the object naming experiment.

This online survey is comprised of three parts:
1. Consent form;
2. Background questionnaire;
3. The main study.

Just for the purpose of the study, please answer all questions in Mandarin Chinese and Simplified Chinese; other languages are not allowed.

Please read the instructions carefully and the mistake examples carefully. No reward will be paid for answers that differ significantly from the experimental requirements.

Theoretically, the whole process will take no more than 40 minutes, but make sure you have enough time to finish this before you start.

If you have any doubts or questions about this study, please send an email to \textit{[email address]}.

You can press [space] to start the experiment whenever you are ready.

\begin{figure}[htbp]
  \centering
  \includegraphics[width=0.49\textwidth]{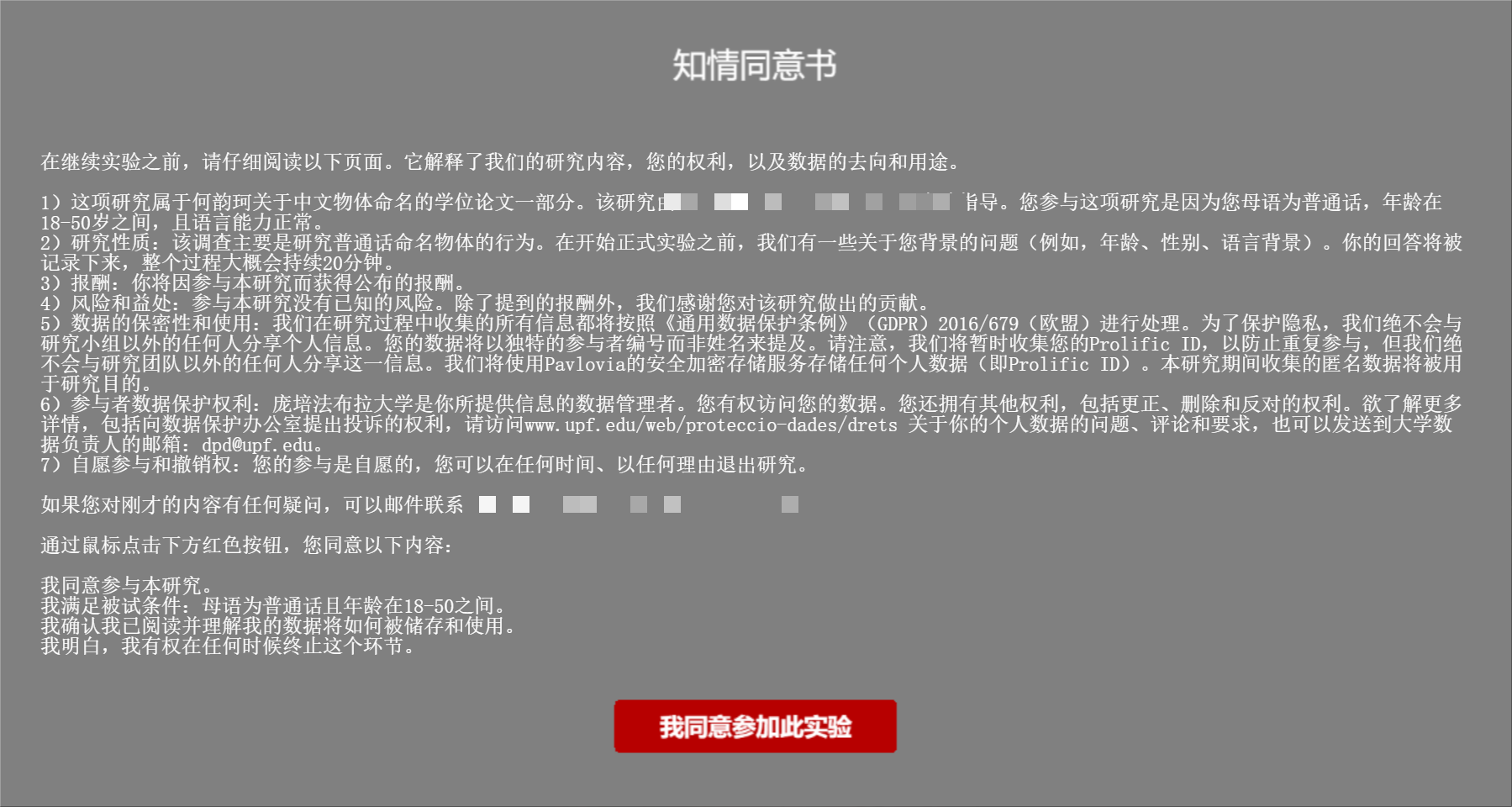} 
  \captionsetup{justification=centering, position=below}
  \caption{Informed Consent Form}
  \label{fig:Informed Consent Form}
\end{figure}

\section*{Translation for Figure \ref{fig:Informed Consent Form}}

Before you proceed with the experiment, please read carefully the following page. It explains our research, your rights, where the data goes, and what it is used for.

\begin{enumerate}
  \item The experiment belongs to \textit{[name]}'s study, supervised by \textit{[name]}.
  You participate in this study because your native language is Mandarin Chinese, age is between 18-50 years old, and you have normal language ability.
  \item Research description: This experiment mainly studies behavior for naming objects in Mandarin Chinese. Before the main experiment, we have some questions about your background (including age, gender, and language backgrounds). Your answer will be recorded, and the process will last approximately 40 minutes.
  \item Reward: You will be paid with the published compensation.
  \item Risks and benefits: Participation in the study entails no unknown risks. Besides the reward mentioned before, we appreciate your contribution to our study.
  \item Privacy: All the information we collect during the course of the research will be processed in accordance with Data Protection Law. In order to safeguard your privacy, we will never share personal information with anyone outside the research team. Your data will be referred to by a unique participant number rather than by name. Please note that we will temporarily collect your Prolific ID to prevent repeated participation; however, we will never share this information with anyone outside the research team. The anonymized data collected during this study will be used for research purposes.
  \item Rights of participants: Pompeu Fabra University is the manager of your data. You have the rights to access your data, including correcting, deleting, and rejecting it. If you want to know more, please access www.upf.edu/web/proteccio-dades/drets. With respect to issues of personal data, you can also send an email to the responsible person of the university: dpd@upf.edu
  \item Voluntary nature of participation: Your participation in this study is on a voluntary basis, and you may withdraw from the study at any time without having to justify why.
\end{enumerate}

By clicking on the red button below, you agree to the following contents:

\begin{itemize}
  \item I agree to participate in this study.
  \item I meet the criteria of participation: my native language is Mandarin Chinese, and my age is between 18-50.
  \item I confirm that I have read all the information above and understand how my data is going to be conserved and used.
  \item I understand that I have the right to terminate this study whenever I want.
\end{itemize}

\begin{figure}[htbp]
  \centering
  \includegraphics[width=0.5\textwidth]{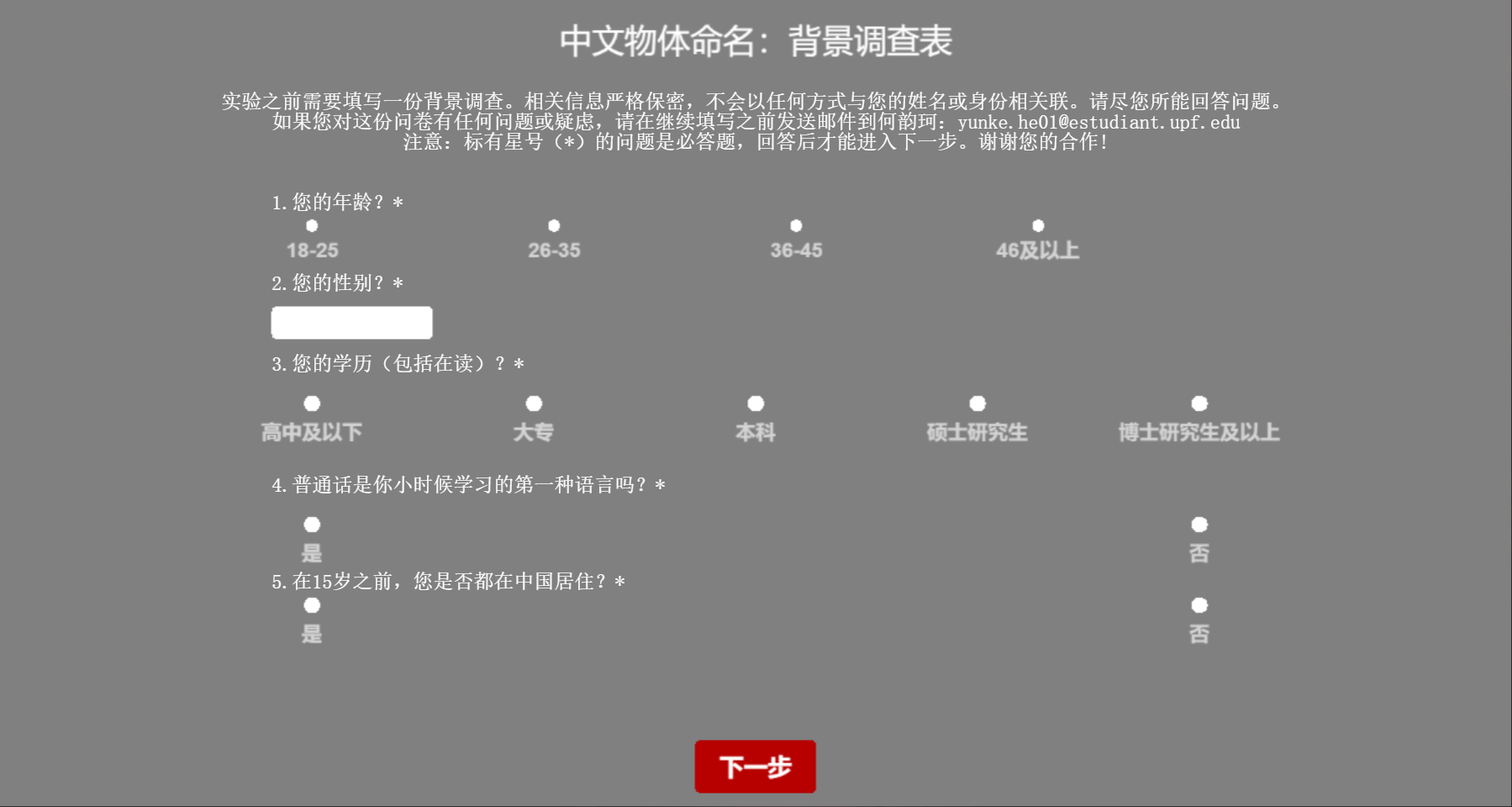} 
  \captionsetup{justification=centering, position=below}
  \caption{Background Survey(A)}
  \label{fig:Background Survey(A)}
\end{figure}

\begin{figure}[htbp]
  \centering
  \includegraphics[width=0.49\textwidth]{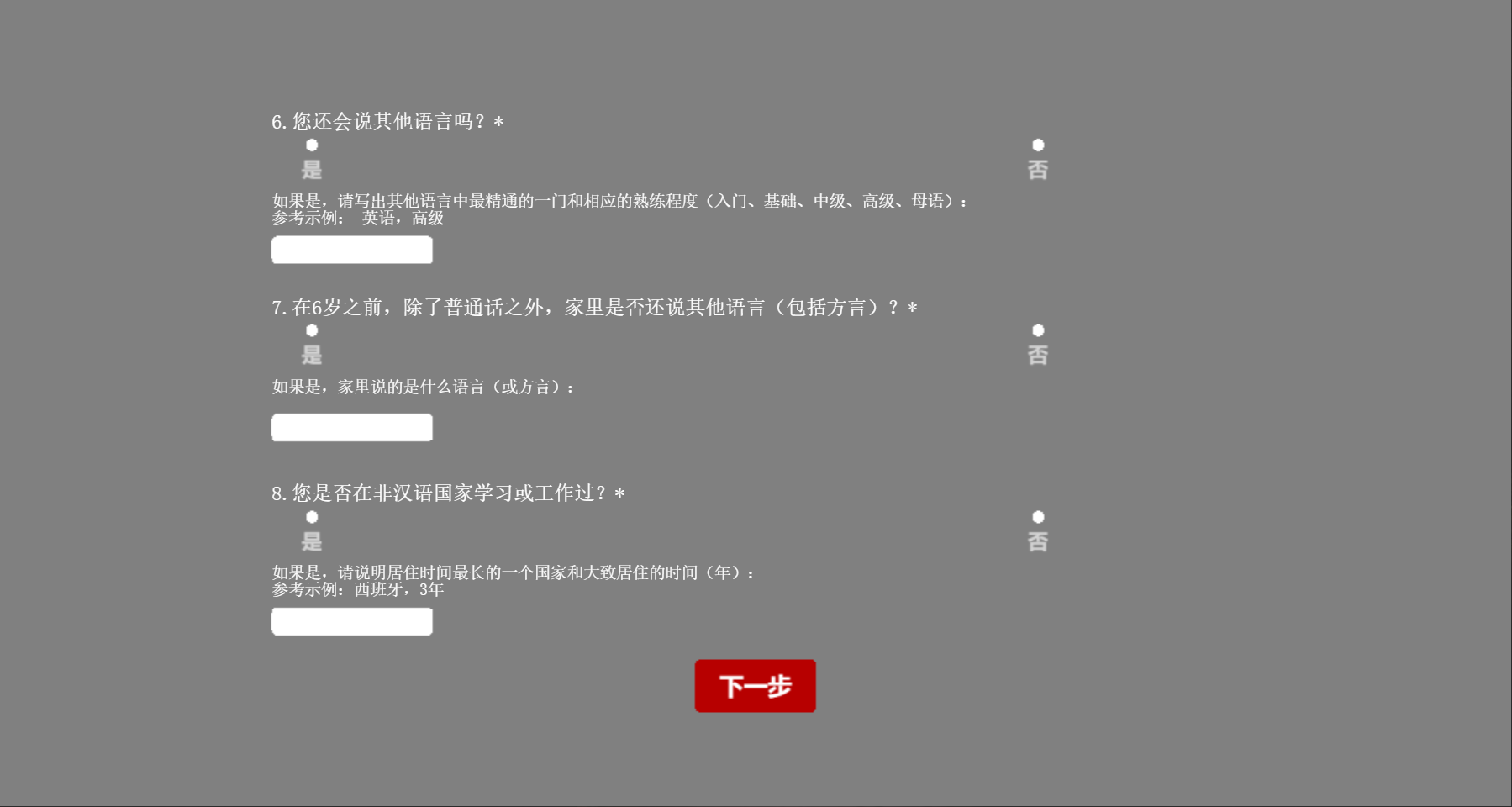} 
  \captionsetup{justification=centering, position=below}
  \caption{Background Survey(B)}
  \label{fig:Background Survey(B)}
\end{figure}

Background survey is translated above in appendix C.

\begin{figure}[htbp]
  \centering
  \includegraphics[width=0.49\textwidth]{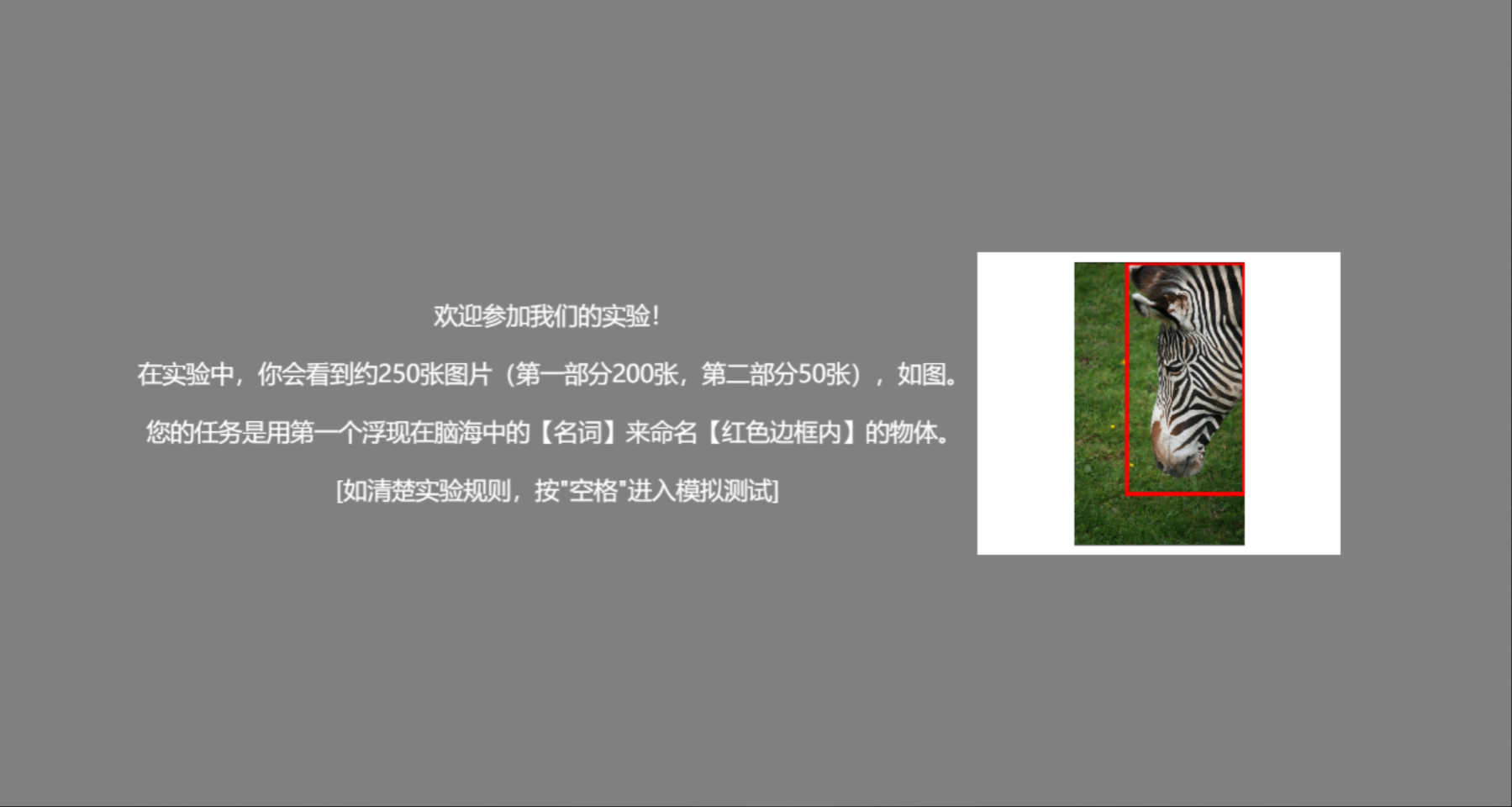} 
  \captionsetup{justification=centering, position=below}
  \caption{Part 1 Introduction}
  \label{fig:Part 1 Introduction}
\end{figure}

\section*{Translation for Figure \ref{fig:Part 1 Introduction}}

Welcome to our study! In the experiment, you will see about 250 images (200 for the first part and 50 for the second part), as shown in the figure. Your task is to name the object in the red bounding box with the first noun that comes to mind.

If you understand the rules, please press [space] to go to the next step.

\begin{figure}[htbp]
  \centering
  \includegraphics[width=0.49\textwidth]{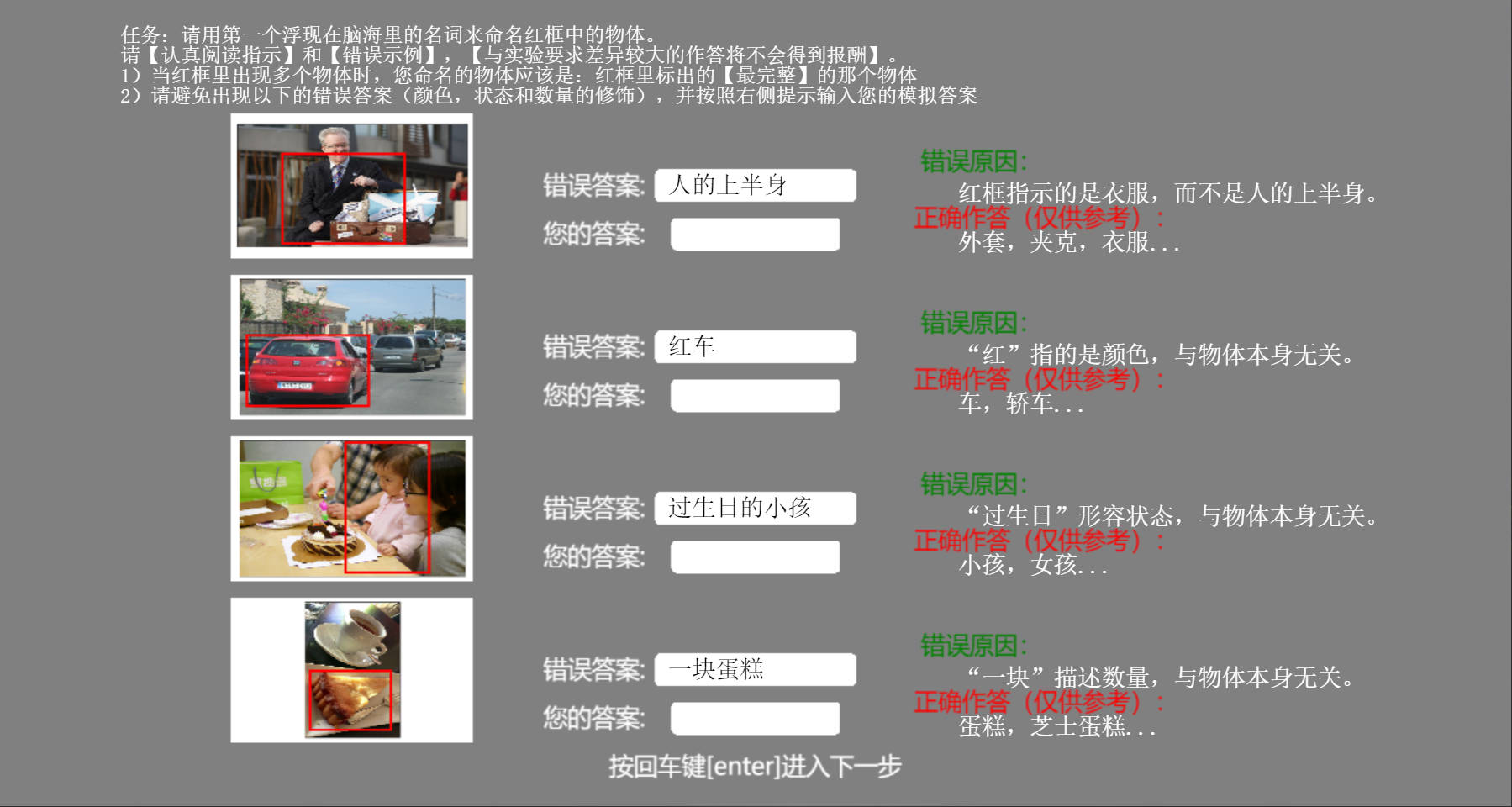} 
  \captionsetup{justification=centering, position=below}
  \caption{Mistakes Exemplified in Part 1}
  \label{fig:Mistakes Exemplified in Part 1}
\end{figure}

\section*{Translation for Figure \ref{fig:Mistakes Exemplified in Part 1}}

Task: Please name the object in the red bounding box with the first noun that came to mind. Please read the instructions carefully and the mistake examples carefully. No reward will be paid for answers that differ significantly from the experimental requirements.

\begin{enumerate}
  \item If multiple objects appear in the red bounding box, the object you should name is the most complete one in the bounding box.
  \item Please try to avoid the mistakes exemplified (modifiers for color, status, and number) and fill in the input box as instructed on the right side.
\end{enumerate}
\textbf{Wrong answer:}The upper part of the human body \\
\textbf{Your answer:} \\
\textbf{Error cause:} The red bounding box indicates the clothes, not the upper part of the human body \\
\textbf{Right answer (just for reference):} jacket, clothes... \\
\textbf{Wrong answer:} red car \\
\textbf{Your answer:} \\
\textbf{Error cause:} "red" refers to the color and has no relation to the object itself \\
\textbf{Right answer (just for reference):} car, taxi...
\textbf{Wrong answer:} the birthday girl \\
\textbf{Your answer:} \\
\textbf{Error cause:} "birthday" refers to the status of the girl and has no relation to the object itself \\
\textbf{Right answer (just for reference):} child, girl...
\textbf{Wrong answer:} a piece of cake \\
\textbf{Your answer:} \\
\textbf{Error cause:} "a piece of" describes the number and has no relation to the object itself \\
\textbf{Right answer (just for reference):} cake, cheesecake

\begin{figure}[htbp]
  \centering
  \includegraphics[width=0.5\textwidth]{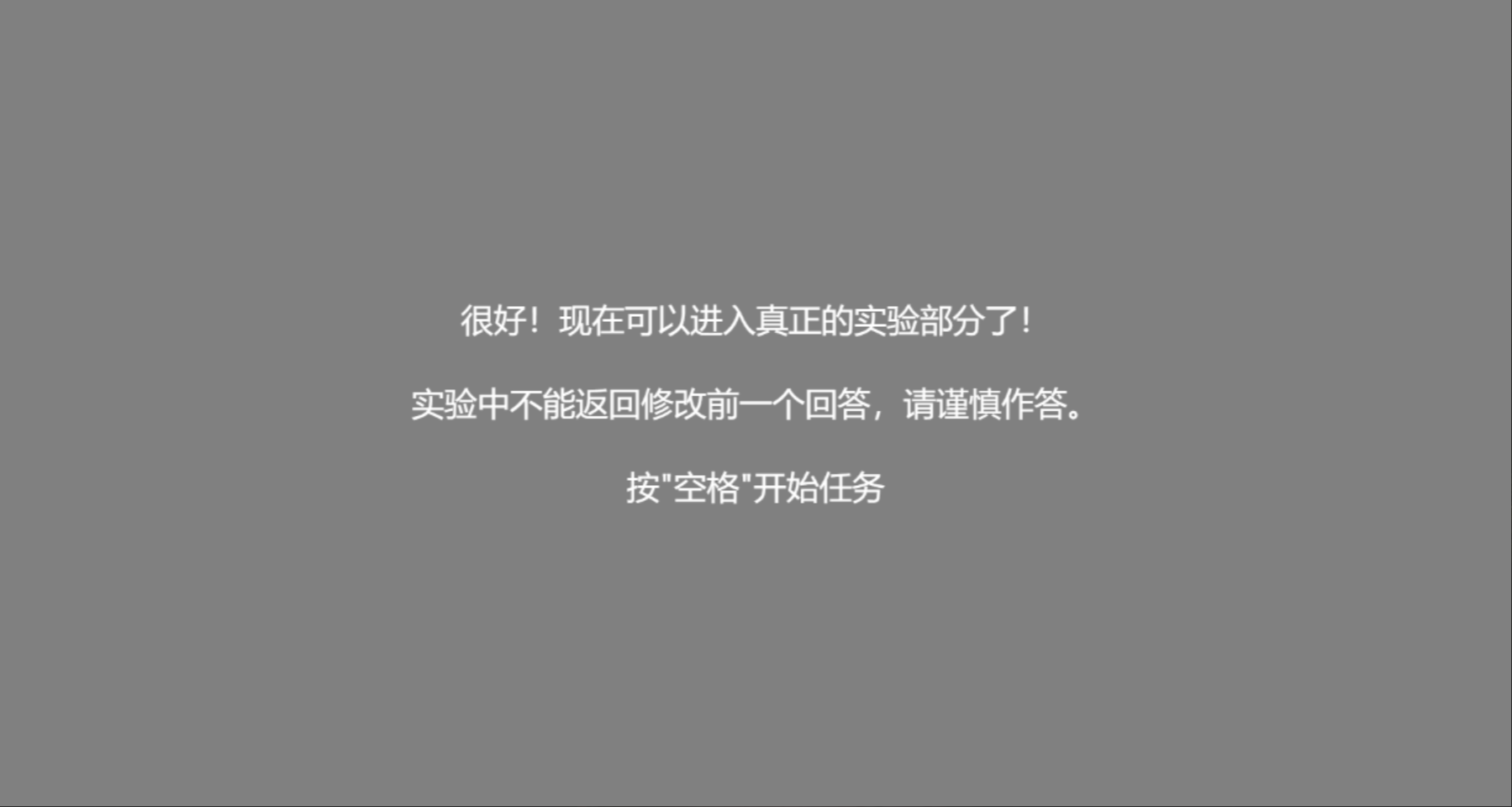} 
  \captionsetup{justification=centering, position=below}
  \caption{Notification for Starting Experiment}
  \label{fig:Notification for Starting Experiment}
\end{figure}

\section*{Translation for Figure \ref{fig:Notification for Starting Experiment}}
Great! Now you can go to the real experiment.

In the experiment you cannot go back to change the previous answer, please answer with caution.

Press [space] to enter the experiment.

\begin{figure}[htbp]
  \centering
  \includegraphics[width=0.49\textwidth]{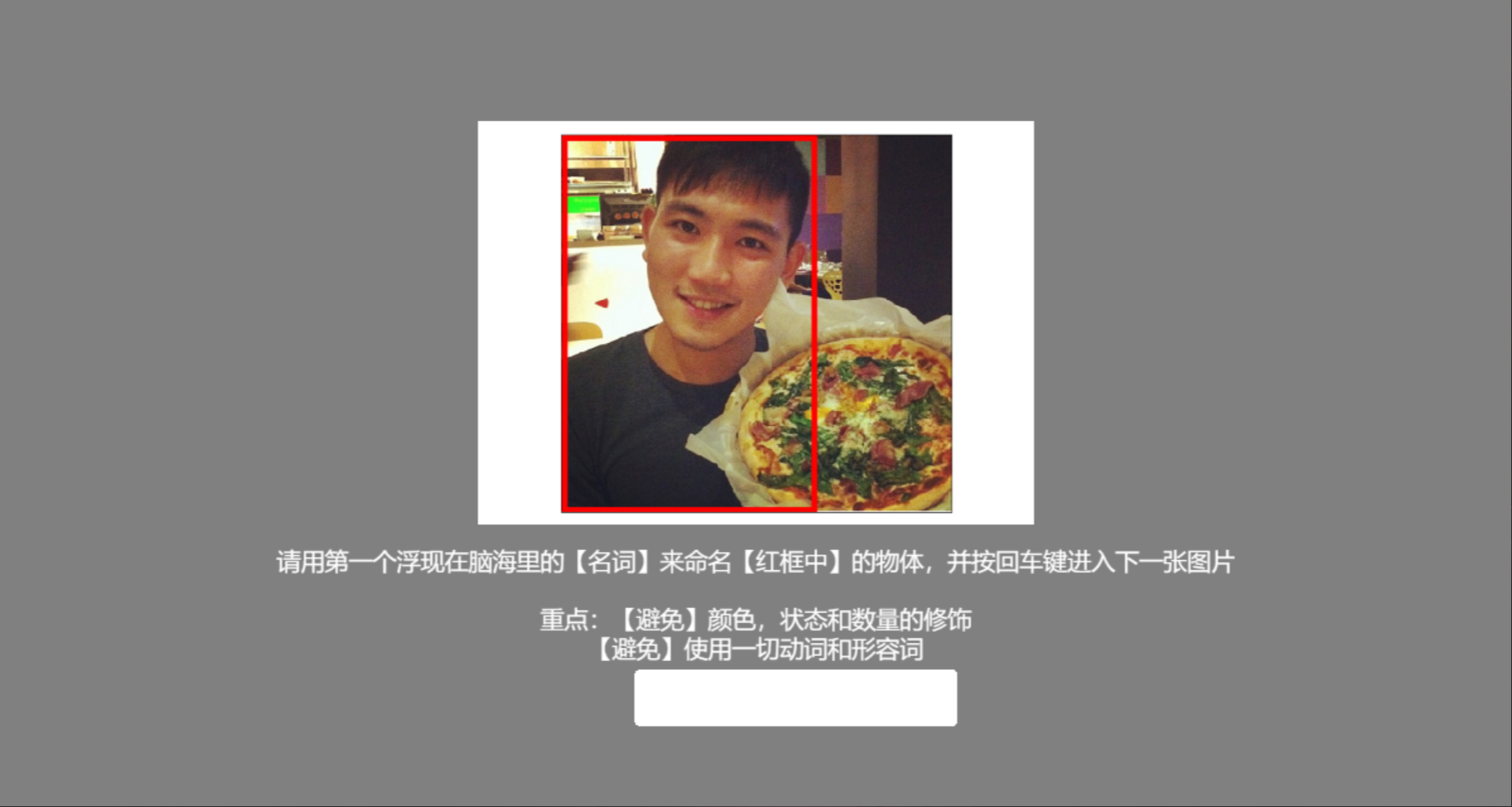} 
  \captionsetup{justification=centering, position=below}
  \caption{Part 1 Object Naming Example}
  \label{fig:Part 1 Object Naming Example}
\end{figure}

\section*{Translation for Figure \ref{fig:Part 1 Object Naming Example}}
Please name the object in the red bounding box with the first noun that came to mind and press [enter] to go to the next image.

Important: avoid modifiers for color, status and number; avoid usage of any verbs and adjectives.

\begin{figure}[htbp]
  \centering
  \includegraphics[width=0.49\textwidth]{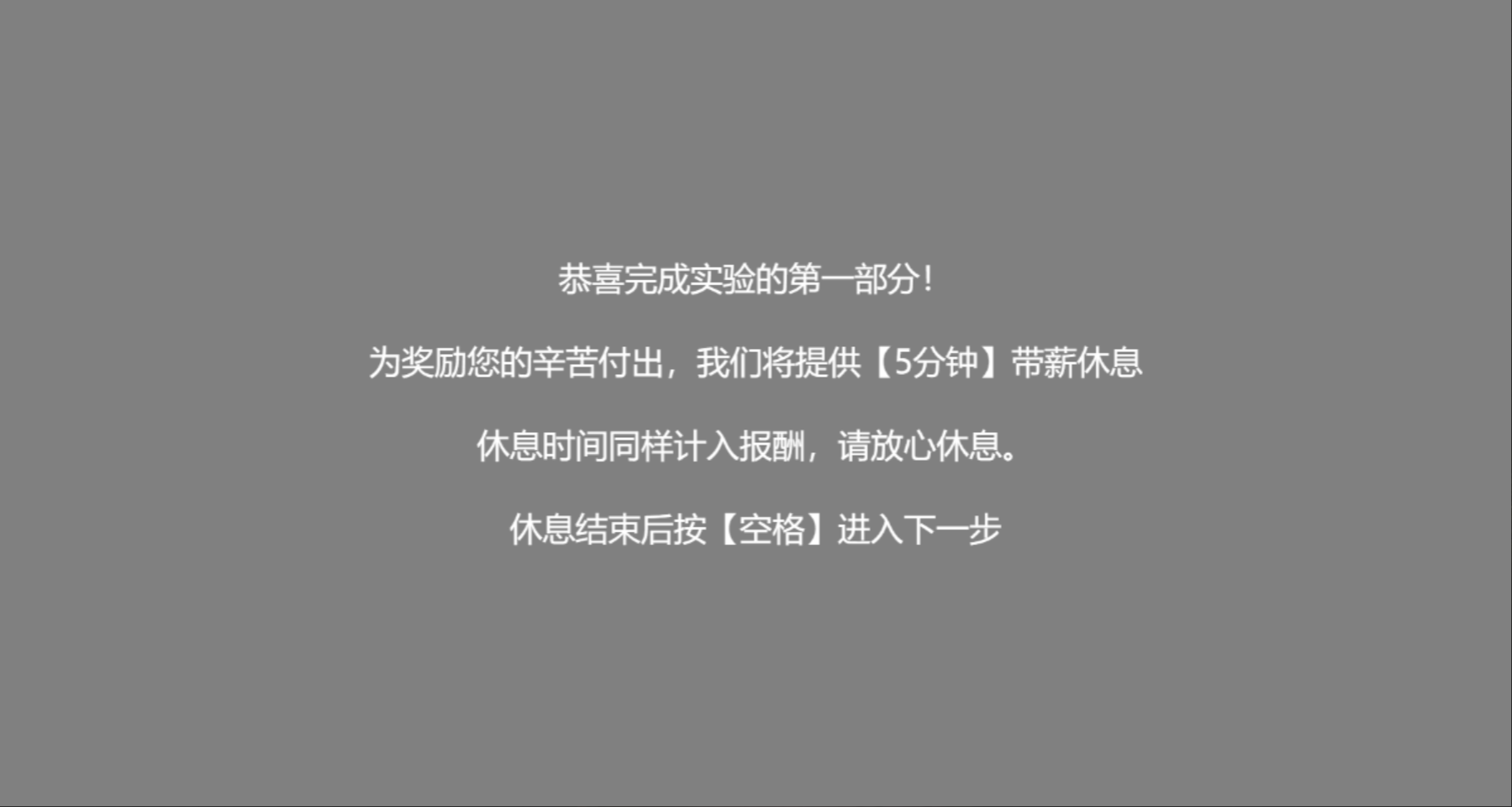} 
  \captionsetup{justification=centering, position=below}
  \caption{5-minute-break between Part 1 and Part 2}
  \label{fig:5-minute-break between Part 1 and Part 2}
\end{figure}

\section*{Translation for Figure \ref{fig:5-minute-break between Part 1 and Part 2}}
Congratulations! You have finished the first part of the experiment!

To reward your hard work, we provide you with five-minute break with compensation included. Please take a rest.

After the break, you can press [enter] to go to the next step.

\begin{figure}[htbp]
  \centering
  \includegraphics[width=0.49\textwidth]{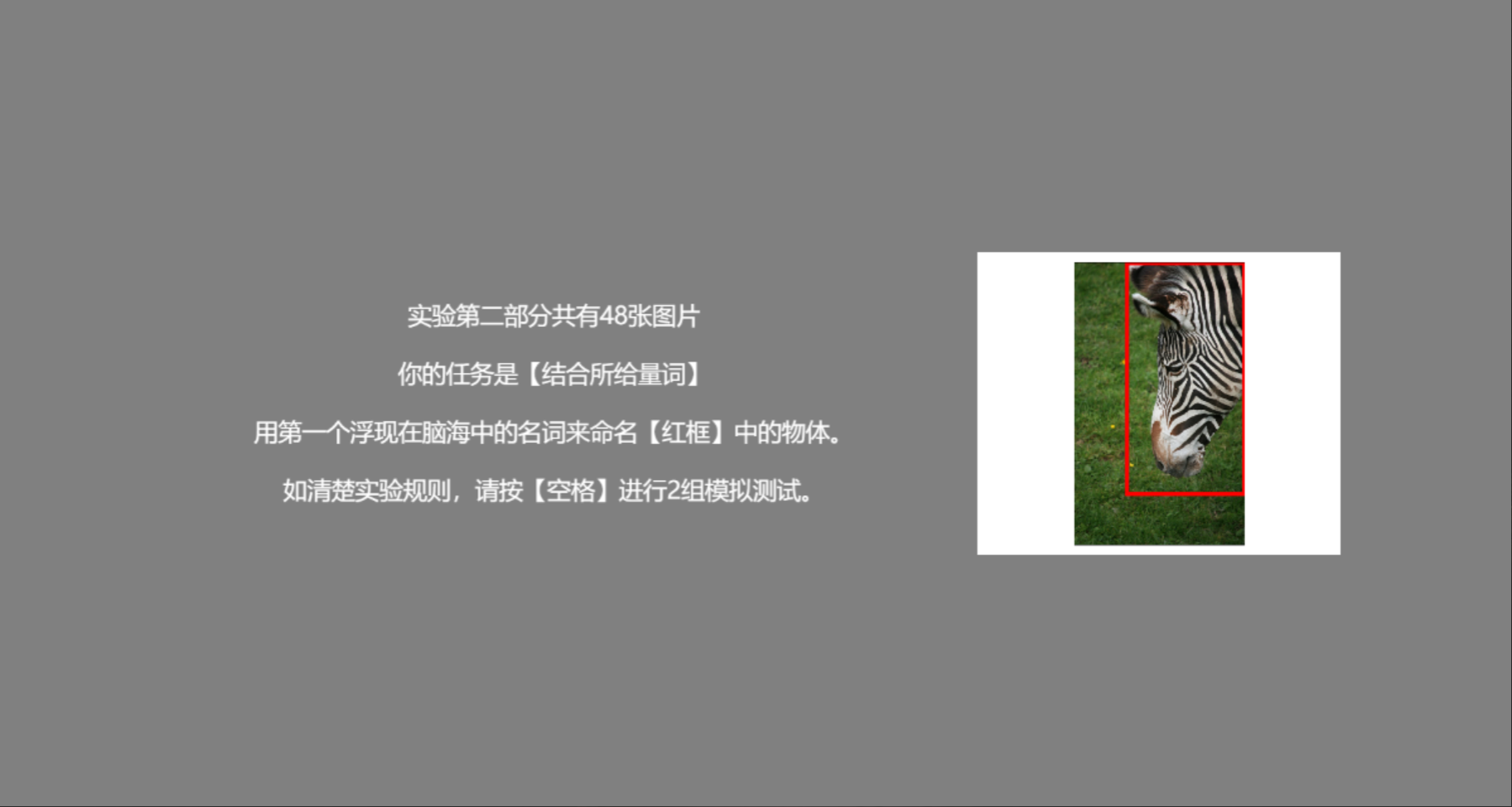} 
  \captionsetup{justification=centering, position=below}
  \caption{Part 2 Introduction}
  \label{fig:Part 2 Introduction}
\end{figure}

\section*{Translation for Figure \ref{fig:Part 2 Introduction}}
The second part of the experiment contains 48 images.

Your task is to name the object in the red bounding box with the first noun that came to mind, combing the classifier we give.

If you understand the rules, please press [space] to go to next step.

\begin{figure}[htbp]
  \centering
  \includegraphics[width=0.5\textwidth]{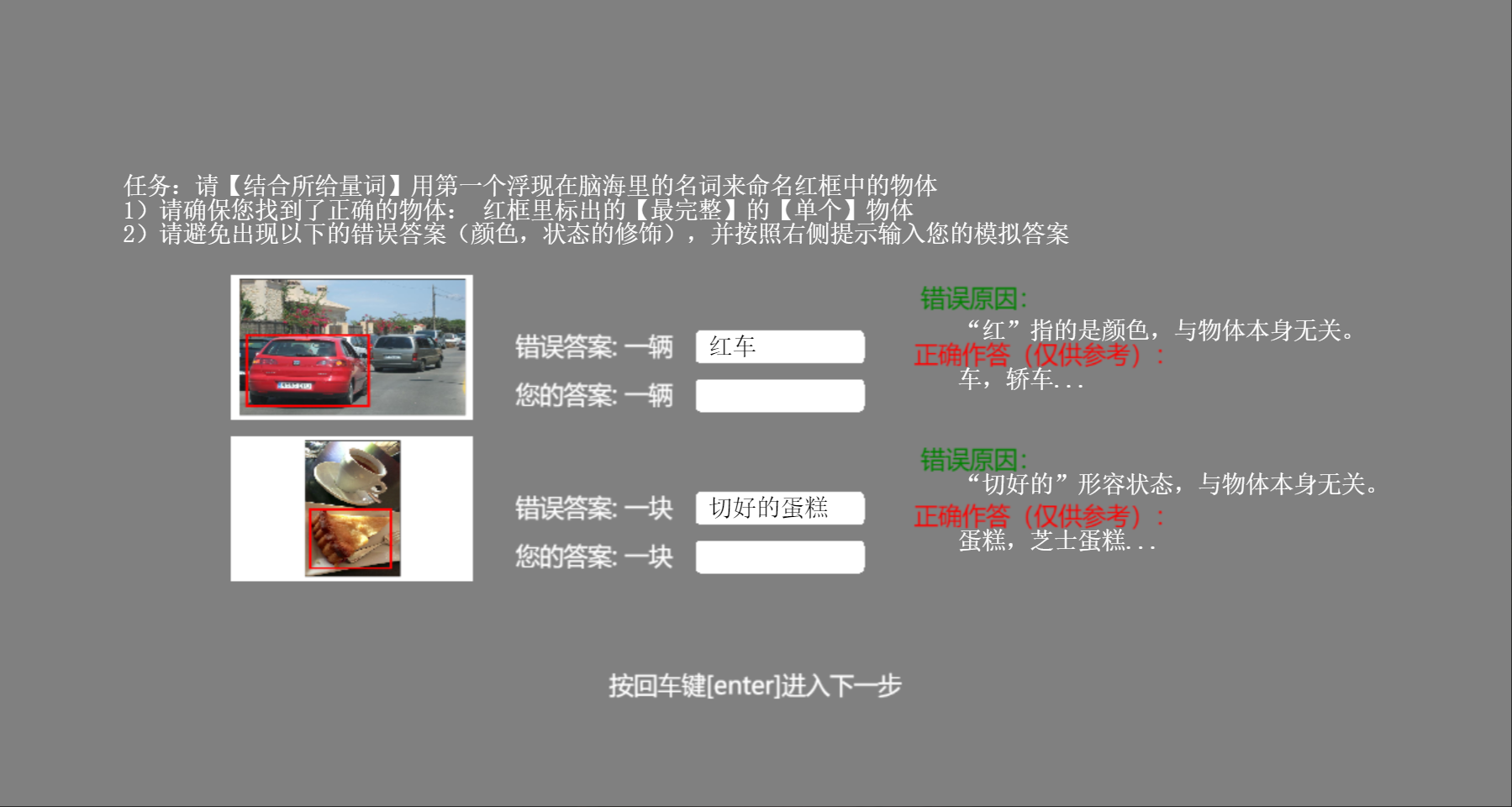} 
  \captionsetup{justification=centering, position=below}
  \caption{Mistakes Exemplified in Part 2}
  \label{fig:Mistakes Exemplified in Part 2}
\end{figure}

\section*{Translation for Figure \ref{fig:Mistakes Exemplified in Part 2}}

Task: please name the object in the red bounding box with the first noun that came to mind, combining the classifier we give.

\begin{enumerate}
\item If multiple objects appear in the red bounding box, the object you should name is the most complete single one in the bounding box.
\item Please try to avoid the mistakes exemplified (modifiers for color and status) and fill in the input box as instructed on the right side.
\end{enumerate}
\textbf{Wrong answer:} one liang of [red car]\\
\textbf{Your answer:} \\
\textbf{Error cause:} the red indicates the color, has no relation to the object itself.\\
\textbf{Right answer (just for reference):} car, taxi...
\textbf{Wrong answer:} one piece of [sliced cake]\\
\textbf{Your answer:} \\
\textbf{Error cause:} sliced indicates the status, has no relation to the object itself.\\
\textbf{Right answer (just for reference):} cake, cheesecake...

\begin{figure}[htbp]
  \centering
  \includegraphics[width=0.5\textwidth]{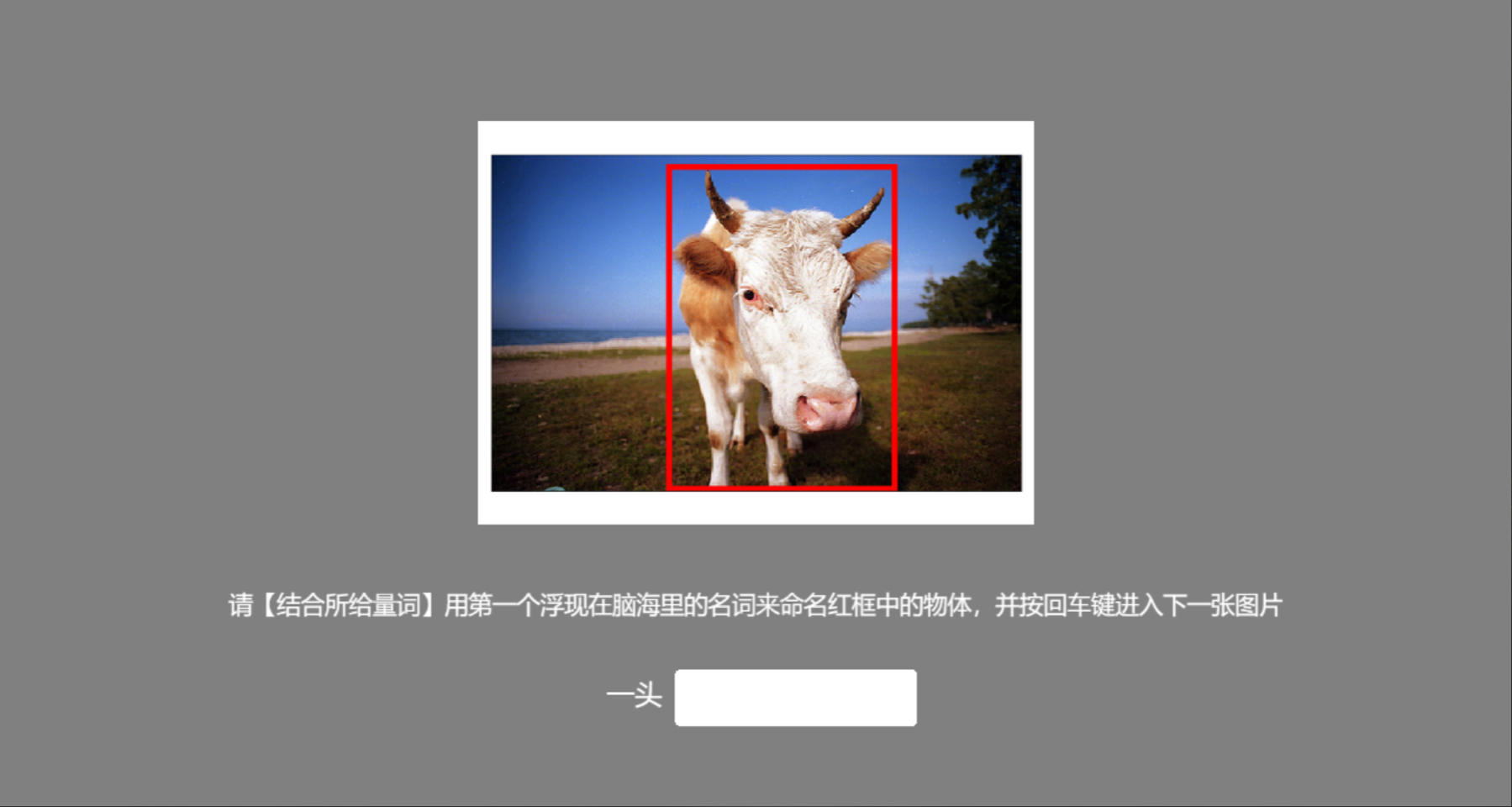} 
  \captionsetup{justification=centering, position=below}
  \caption{Part 2 Object Naming with Classifier Example}
  \label{fig:Part 2 Object Naming with Classifier Example}
\end{figure}

\begin{figure}[htbp]
  \centering 
  \includegraphics[width=0.49\textwidth]{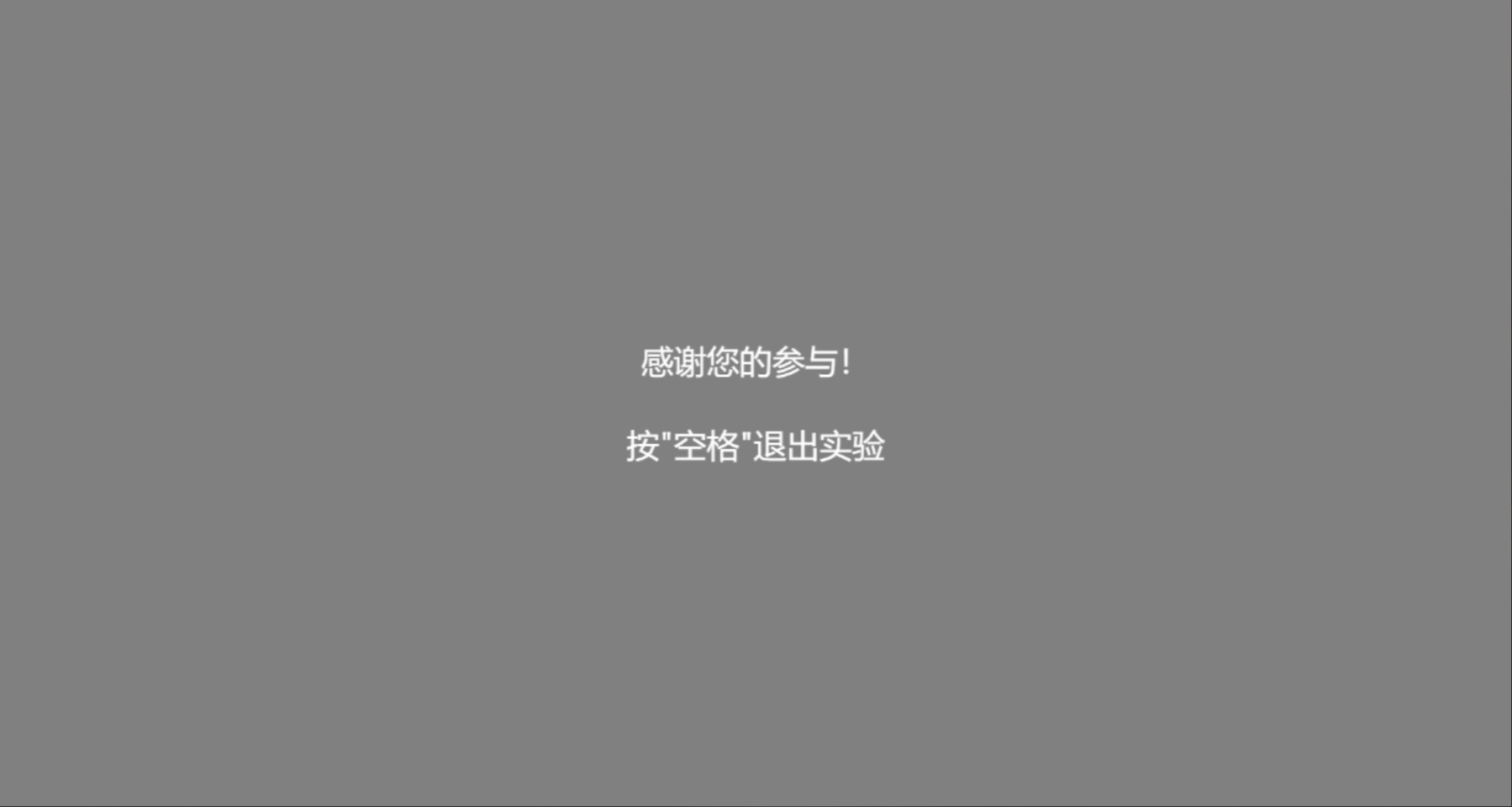} 
  \captionsetup{justification=centering, position=below}
  \caption{End}
  \label{fig:End}
\end{figure}

\section*{Translation for Figure \ref{fig:Part 2 Object Naming with Classifier Example}}
please name the object in the red bounding box with the first noun that came to mind, combing the classifier we give, and press [enter] to go to the next image.

\section*{Translation for Figure \ref{fig:End}}
Thanks a lot for your participation!

Press [space] to exit.

\clearpage

\end{document}